\PassOptionsToPackage{dvipsnames}{xcolor}
\documentclass[sigconf]{acmart} %
\AtBeginDocument{%
  }

\usepackage{soul}
\usepackage[breakable,skins]{tcolorbox}
\usepackage{adjustbox}
\usepackage[capitalise]{cleveref}
\Crefname{figure}{Fig.}{Figs.}
\Crefname{table}{Tab.}{Tabs.} 
\Crefname{section}{Sec.}{Secs.} 
\Crefname{subsection}{Sec.}{Secs.}
\Crefname{equation}{Eq.}{Eqs.}
\usepackage[T1]{fontenc}
\usepackage{newtxtext}
\usepackage{bm}
\usepackage{amsmath}
\DeclareMathAlphabet{\mymathbb}{U}{BOONDOX-ds}{m}{n}  %
\usepackage{xspace} 
\usepackage{multirow} 
\usepackage{tabularx}
\usepackage{marvosym}
\usepackage{colortbl}
\usepackage{pifont}
\usepackage[normalem]{ulem}
\usepackage{enumitem}
\usepackage{graphicx}
\graphicspath{{figures/}}
\usepackage{relsize}
\usepackage{nicematrix}
\usepackage{wrapfig}
\usepackage{makecell}
\usepackage{cleveref}
\usepackage{adjustbox}
\usepackage{subfig}
\usepackage{balance}

\definecolor{goodgreen}{rgb}{0.0, 0.56, 0.0}
\definecolor{badgray}{HTML}{666666}
\definecolor{highlight}{rgb}{0.98, 0.94, 1.0}
\definecolor{ourpurple}{RGB}{225, 213, 231}
\definecolor{ourred}{RGB}{234, 107, 102}
\definecolor{ouryellow}{RGB}{230, 195, 92}
\definecolor{ourorange}{RGB}{180, 101, 4}
\definecolor{ourgreen}{RGB}{213, 232, 212}
\definecolor{ourblue}{RGB}{108, 142, 191}
\definecolor{ourgray}{RGB}{102, 102, 102}

\let\titleold\title
\renewcommand{\title}[1]{\titleold{#1}\newcommand{\thetitle}{#1}}
\def\maketitlesupplementary
   {
   \newpage
       \twocolumn[
        \centering
        \Large
        \textbf{\thetitle}\\
        \vspace{0.5em}Supplementary Material \\
        \vspace{1.0em}
       ] %
   }

\usepackage{calc}

\newlength{\DepthReference}
\newlength{\HeightReference}
\newlength{\Width}%

\newcommand{\MyColorBox}[2][red]%
{%
    \settowidth{\Width}{#2}%
    \fcolorbox{#1}{white}%
    {%
        \raisebox{-\DepthReference}%
        {%
                \parbox[b][\HeightReference+\DepthReference][c]{\Width}{\centering\textcolor{#1}{#2}}%
        }%
    }%
}

\newcommand{\ie}{\textit{i.e.}\xspace}
\newcommand{\eg}{\textit{e.g.}\xspace}
\newcommand{\cmark}{\textcolor{ForestGreen}{\ding{51}}}
\newcommand{\xmark}{\textcolor{red}{\ding{55}}}
\newcommand{\method}{\textsc{dysco}\xspace}
\newcommand{\methodaf}{\textsc{dysco-lf}\xspace}
\newcommand{\multihead}{\textsc{MhOM}\xspace}
\newcommand{\multiheadacro}{Multi-head Orchestrator Module\xspace}
\newcommand{\plusours}{ \ \ \ \ $\mathbf{+}$ \ $\bm{\mathbb{S}}$}

\copyrightyear{2025}
\acmYear{2025}
\setcopyright{cc}
\setcctype{by-nc-sa}
\acmConference[MM '25]{Proceedings of the 33rd ACM International Conference on Multimedia}{October 27--31, 2025}{Dublin, Ireland}
\acmBooktitle{Proceedings of the 33rd ACM International Conference on Multimedia (MM '25), October 27--31, 2025, Dublin, Ireland}
\acmDOI{10.1145/3746027.3754770}
\acmISBN{979-8-4007-2035-2/2025/10}

\acmSubmissionID{537}

\begin{document}

\title{Dynamic Scoring with Enhanced Semantics for Training-Free Human-Object Interaction Detection}

\author{Francesco Tonini}
\orcid{0000-0002-1938-3449}
\affiliation{%
  \institution{University of Trento}
  \city{Trento}
  \country{Italy}
}
\affiliation{%
  \institution{Fondazione Bruno Kessler}
  \city{Trento}
  \country{Italy}
}
\email{francesco.tonini@unitn.it}

\author{Lorenzo Vaquero}
\orcid{0000-0002-1874-3078}
\affiliation{%
  \institution{Fondazione Bruno Kessler}
  \city{Trento}
  \country{Italy}
}
\email{lvaquerootal@fbk.eu}

\author{Alessandro Conti}
\orcid{0000-0002-3044-1320}
\affiliation{%
  \institution{University of Trento}
  \city{Trento}
  \country{Italy}
}
\email{alessandro.conti-1@unitn.it}

\author{Cigdem Beyan}
\orcid{0000-0002-9583-0087}
\affiliation{%
  \department{Department of Computer Science}
  \institution{University of Verona}
  \city{Verona}
  \country{Italy}
}
\email{cigdem.beyan@univr.it}

\author{Elisa Ricci}
\orcid{0000-0002-0228-1147}
\affiliation{%
  \institution{University of Trento}
  \city{Trento}
  \country{Italy}
}
\affiliation{%
  \institution{Fondazione Bruno Kessler}
  \city{Trento}
  \country{Italy}
}
\email{e.ricci@unitn.it}

\begin{abstract}
Human-Object Interaction (HOI) detection aims to identify humans and objects within images and interpret their interactions. Existing HOI methods rely heavily on large datasets with manual annotations to learn interactions from visual cues. These annotations are labor-intensive to create, prone to inconsistency, and limit scalability to new domains and rare interactions. We argue that recent advances in Vision-Language Models (VLMs) offer untapped potential, particularly in enhancing interaction representation. While prior work has injected such potential and even proposed training-free methods, there remain key gaps. Consequently, we propose a novel training-free HOI detection framework for \textbf{Dy}namic \textbf{Sco}ring with enhanced semantics (\method) that effectively utilizes textual and visual interaction representations within a multimodal registry, enabling robust and nuanced interaction understanding. This registry incorporates a small set of visual cues and uses innovative interaction signatures to improve the semantic alignment of verbs, facilitating effective generalization to rare interactions.
Additionally, we propose a unique multi-head attention mechanism that adaptively weights the contributions of the visual and textual features.
Experimental results demonstrate that our \method surpasses training-free state-of-the-art models and is competitive with training-based approaches, particularly excelling in rare interactions. Code is available at \url{https://github.com/francescotonini/dysco}.
\end{abstract}




\maketitle

\section{Introduction} \label{sec:intro}

\begin{figure}[!t]
\centering
\includegraphics[width=0.85\linewidth]{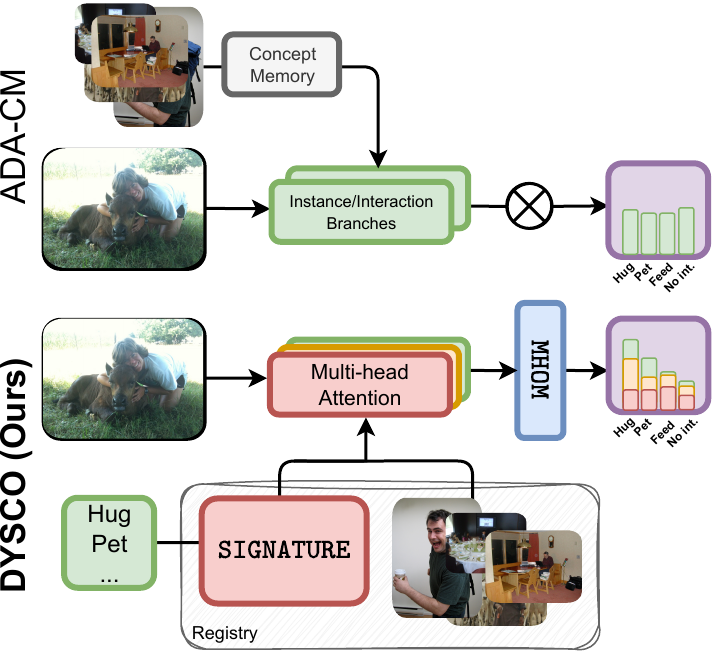}
\caption{We introduce \method, a training-free Human-Object Interaction (HOI) detector that leverages a multimodal registry enriched with fine-grained interaction representations denoted as \textit{signatures}. Unlike ADA-CM~\cite{Lei2023}, the only existing training-free model,
which relies mainly on visual features, \method integrates multimodal data and adaptively reweights multimodal head scores based on the unique characteristics of each test sample, improving the detection of complex interactions.}
\Description{Previous training-free HOI detectors rely mainly on visual features. Our proposal, \method, leverages a multimodal registry enriched with fine-grained interaction representations and leverages a multi-head attention for information fusion.}
\label{fig:teaser}
\vspace{-1.5em}
\end{figure}

Human-Object Interaction (HOI) detection focuses on accurately identifying humans and objects within images and understanding the interactions between them. Formally, given an image, HOI seeks to locate human-object pairs and identify their interactions as a set of \texttt{<human, verb, object>} triplets. This capability is highly valuable for various downstream applications, such as image and video captioning~\cite{9583890,Seo_2022_CVPR}, visual surveillance~\cite{li2024ripple}, and autonomous driving~\cite{chen2024asynchronous}, as it greatly improves perception and understanding within automated systems.

Recently, Vision Transformers~\cite{vaswani2017attention}, particularly DETR~\cite{Carion2020}, have brought significant advancements to HOI detection.
Two-stage approaches utilize DETR to first localize humans and objects and later use the features from these detections to classify interactions~\cite{Lei2023,Mao2023,Zhang2022,Kim2020b,Ulutan2020,Gao2020,Zhang2021}. Alternatively, one-stage methods fine-tune DETR-based architectures to directly predict HOI triplets from the input image in a unified, end-to-end process~\cite{Kim2022,Tu2022,Zhong2022}.
Another important development is the incorporation of Vision-Language Models (VLMs), particularly CLIP~\cite{Radford2021}, which has shown strong potential for improving HOI performance.
For instance, Cao et al.~\cite{Cao2023} leverage VLMs to compute the similarity between textual descriptions and detection proposals, whereas~\cite{Wang2022b} learns textual prompts to better match the feature space of HOI.

Despite improvements, most methods, including those relying on VLMs~\cite{Ning2023,Mao2023,wang2024bilateral} remain fully supervised 
and depend on a massive amount of manual annotations at the HOI instance level.
Annotating HOI pairs, however, is not only highly labor-intensive and time-consuming but also subjective, as it often depends on individual interpretation, and leads to inconsistencies. This process further exacerbates data scarcity, particularly when applied to new domains or situations where annotated data is limited or unavailable. Furthermore, the inherent combinatorial nature of HOIs further complicates the task, especially when dealing with rare interactions in long-tailed distributions. 
Indeed, many methods suffer from poor performance on rare interactions~\cite{Kim2021,Tamura2021,Li2022a}.
Furthermore, training or fine-tuning HOI detectors is computationally demanding. Two-stage methods require an exhaustive combination of instance-level features to predict relationships, while one-stage detectors
encounter difficulties due to their heavy dependence on transformers.

The challenges mentioned above have been addressed to some extent in ADA-CM~\cite{Lei2023}, which presents the first and only \textit{training-free} HOI detection pipeline, with a primary focus on achieving on-par performance for both rare and non-rare classes. ADA-CM utilizes DETR to generate \texttt{<human, object>} pairs and extracts 
instance-centric 
(related to pose and orientation) and interaction-aware features (referring to contextual information) for each proposal.
It builds a memory system driven by visual features, all encoded by CLIP~\cite{Radford2021}, expecting that these features enable the model to leverage visual and text commonsense to capture potential co-occurrences and relationships between objects and interactions. However, this may often lead to incorrect associations between text and vision, as the HOI task is considerably more complex than image classification. HOI tasks include not only objects but also actions, which require richer semantic understanding. Also, ADA-CM overlooks the contributions of instance-centric and interaction-aware features, which may not be equal at all times. The contributions of these features could be dynamically adjusted in a training-free manner, as they may vary from verb to verb (see \cref{sec:qualitatives}).

In this paper, we introduce \method, a novel \textit{training-free} HOI detector that follows a two-stage approach, leveraging human and object proposals from DETR~\cite{Carion2020}.
Recognizing that VLMs' textual encoders primarily capture nouns and adjectives but struggle with verbs due to their limited semantic information~\cite{Bhalla2024,Momeni2023},
we propose an effective strategy for improving verb comprehension and interaction representation without requiring fine-tuning and/or adaptation. This technique, referred to as \textit{interaction signature generation}, extracts \textit{action-semantic tokens} that enrich the interaction representation, improving its classification capabilities.
We formulate the HOI task as a multi-head attention process (see~\cref{fig:teaser}), 
where each head independently processes distinct visual and textual features, dynamically contributing to the final prediction. This process includes a negative bias that deals with visually similar interactions.
This provides a viable, training-free solution that dynamically emphasizes both fine-grained and contextual multimodal information as required.

Experimental analysis on standard HOI datasets confirms that \method outperforms state-of-the-art (SOTA) training-free methods as well as several training-based approaches, particularly on rare classes. Furthermore, the ablation study confirms the importance of each component. We also demonstrate the universality of our method, as altering the VLM backbone (e.g., by scaling up or employing extended versions) consistently surpasses prior approaches~\cite{Lei2023}.

Our contributions can be summarized as follows: 
\begin{itemize}[leftmargin=*]
    \item We propose \method, a novel training-free HOI detector that effectively harnesses both rich textual and visual information, enabling robust and accurate human-object interaction detection.
    \item \method introduces an innovative method for generating interaction signatures that remarkably improves the semantic alignment between interaction representations and visual features.
    \item We successfully cast the HOI prediction task as a training-free multi-head attention process, enabling, for the first time in HOI, dynamic reweighting and specialization of multimodal heads to improve VLMs' predictions and visually similar interactions.
    \item \method establishes a new SOTA for training-free HOI detection and is competitive with training-based approaches (\cref{sec:SOTAcomp}). We provide a comprehensive analysis of its components and performance against prior arts (\cref{sec:SOTAcomp} and \cref{sec:ablation}), even in the absence of manually-curated annotations (\cref{sec:pseudolabels}).
\end{itemize}

\section{Related work} \label{sec:related}

\textbf{Human-object interaction detection.}
One-stage HOI detectors treat the task as a set prediction problem, simultaneously performing object detection, object association, and interaction classification.
Earlier versions of such methods used bounding-box unions~\cite{Kim2020a} and interaction points~\cite{Liao2020,Wang2020} to capture interaction regions.
However, more recent one-stage methods follow a DETR-like~\cite{Carion2020} architecture and leverage learnable queries, which are fed to a Transformer decoder to predict the triplets.
These one-stage approaches can be further categorized into single-branch and two-branch methods: single-branch methods use a single decoder to predict the HOI instances~\cite{Kim2022,Tu2022,Zhong2022}, while two-branch methods employ one decoder to detect human-object pairs and another to classify their interactions~\cite{Kim2021,Kim2023,Zhou2022}. Although one-stage methods perform well in fully-supervised settings, they are often computationally intensive, slow to converge, and unsuitable for training-free scenarios, as their joint localization strategy performs best only when relying on large labeled datasets and lacks the adaptability required for new action-object combinations without retraining. For these reasons, two-stage methods are generally preferred for tasks that require broad generalization and zero-shot capabilities.

\begin{figure*}[!ht]
\centering
\includegraphics[width=0.9\linewidth]{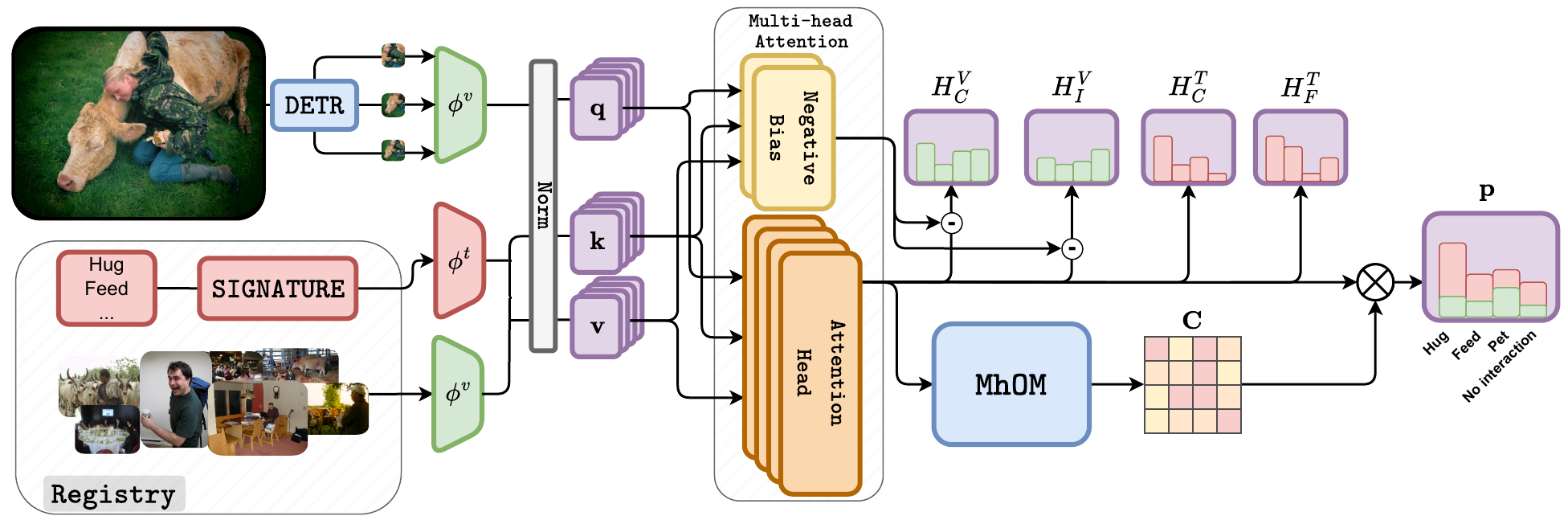}
\caption{Our \method. We begin by generating \MyColorBox[ourred]{novel interaction signatures}, which enhance the semantic information of textual categories. We also utilize a \MyColorBox[ourblue]{object detector} to identify humans and objects in the image and extract visual features from the crops of the detected human, object, and their union bounding box. A set of \MyColorBox[ourorange]{attention heads} then processes the features of the test sample alongside those of the \MyColorBox[ourgray]{registry} of interaction signatures and annotated images from the dataset. Furthermore, \MyColorBox[ouryellow]{negative biases} are attached to visual heads to improve the performance of predicting visually similar interactions.
Finally, we adaptively reweight the contribution of each attention head using our \MyColorBox[ourblue]{\multiheadacro}, selectively emphasizing heads that provide interaction-relevant information on a per-interaction basis.}
\Description{Architecture of \method, that leverages text-based interaction signatures and visual features to yield precise HOI predictions.}
\label{fig:method}
\end{figure*}

Two-stage HOI detectors typically begin by using pre-trained object detectors~\cite{Carion2020,Ren2017} to generate human and object proposals. They then enumerate potential human-object pairs and apply various techniques, such as visual attention~\cite{Zhang2022}, co-occurrence priors~\cite{Kim2020b}, spatial features~\cite{Ulutan2020,Gao2020,Zhang2021}, and pose features~\cite{Li2022a,Li2020,Wu2024b}, to refine the features for the interaction classifier. The classifier subsequently generates predictions through relation modeling strategies, such as message-passing in graph structures~\cite{Ulutan2020,Zhang2021,Liu2020b} or multi-stream fusion~\cite{Chao2018,Gao2020}.
Although these techniques enhance detection performance, both one-stage and two-stage approaches significantly underperform on rare classes, owing to the long-tailed distribution of HOI training data~\cite{Lei2023,Mao2023}. \\

\vspace{-0.5em}
\noindent \textbf{Vision-Language Models for HOI.}
The extensive and diverse data used to train VLMs (i.e., CLIP \cite{Radford2021}) equips them with a deep understanding of the real world, which can be leveraged across a wide range of tasks~\cite{Radford2021,Li2022c,Zhang2024}.
These models can support HOI detection in various ways: by replacing the image feature extractor with a VLM image encoder~\cite{Park2023,Ning2023} to generate high-quality features that are more resilient to out-of-domain samples, by prompting the model to capture additional HOI cues from the scene~\cite{Mao2023,Cao2023,Luo2024,Wu2024} for later integration into the architecture, or by distilling knowledge from large models to enhance performance in supervised learning~\cite{Liao2022,Wu2023,Wang2022b,Zhao2023,wang2024bilateral}. 
However, all these methods share a common limitation: they require fine-tuning and/or adaptation of
the VLM or other components of the HOI architecture on the downstream task, which means they are still affected by the long-tailed distribution of HOI training data~\cite{Lei2023,Mao2023}. 
Furthermore, as noted in~\cite{xie2025relationlmm}, even multimodal large language models without HOI supervision fail to achieve SOTA performance in HOI tasks.
ADA-CM~\cite{Lei2023}, has recently addressed this challenge by introducing a \textit{training-free} alternative for HOI detection.
However, their method is constrained for two major reasons. First, they rely solely on visual representations to build a memory, assuming that both visual and textual commonsense can be naturally derived from CLIP~\cite{Radford2021}. However, we empirically show that this assumption does not hold in all cases (e.g., rare class), and text can further enhance the visual embeddings by considering a more semantically meaningful region. Second, ADA-CM treats the features extracted from e.g., the pose or orientation of DETR's~\cite{Carion2020} detection proposals, as well as environmental and contextual information, with equal importance. However, this approach is flawed, as it is incorrect to assume that these features contribute equally at all times, given the high variability in HOI instances. Consequently, we present \method, which addresses all these issues by implementing a novel signature interaction generation and a multi-head attention process that allows dynamic reweighting of visual and textual features. \\

\vspace{-0.5em}
\noindent \textbf{Semantic representation understanding.}
Recently, there has been a shift in the epistemological perspective of machine learning, transitioning from merely extracting labels from images to interpreting these labels through specialized encoders~\cite{Pearl2021}.
To improve the coupling of data representations, several efforts have focused on defining a more appropriate shared space, often employing discrete key-value bottlenecks~\cite{Trauble2023}.
Some of these methods even achieve this without training~\cite{Norelli2023}, drawing inspiration from compressed sensing algorithms~\cite{Wang2016,Locatello2018}.
Other approaches pursue alignment by adopting a relative representation technique that aligns the latent spaces of a single model trained on different domains~\cite{Moschella2023}, with advancements like \cite{Maiorca2023} using closed-form solutions to relax some of the constraints.
Nevertheless, a primary limitation of these methods is their difficulty in aligning embeddings with limited semantic information, such as verbs processed by CLIP’s text encoder~\cite{Radford2021}.

To quantify the semantic content within dense representations, the linear representation hypothesis proposes that semantic concepts are linearly organized in a model's latent space~\cite{Park2024}. 
This structure enables a translation between modality-specific dense embeddings and sparse semantic representations.
This representations can be achieved through concept bottleneck models~\cite{Koh2020}, mechanistic interpretability~\cite{Fel2023}, or disentangled representation learning~\cite{Hsu2023}.
As these methods typically depend on qualitative visualizations or predefined concept sets, there has been a late rise in post-hoc approaches~\cite{Bhalla2024}.
However, a common limitation among these methods is their reduction of each concept to a single, fixed representation, which can limit its expressiveness and generalizability, as displayed in~\cref{fig:signature}.
To address this, we propose a novel signature generation process designed to capture the complex and stratified manifolds of more nuanced concepts, such as interactions.

\section{Method}
\label{sec:method}

\method is a training-free approach that leverages rich textual and visual information for robust HOI detection.
As illustrated in \cref{fig:method}, the process begins with a novel interaction signature generation (\cref{sec:prompts_expansion}), which enhances the semantic information of the textual categories.
Next, \method identifies all humans and objects in the scene and pairs them exhaustively (\cref{sec:detector}), extracting their visual features.
Finally, it classifies the various interactions by framing the task as a multi-head attention process with dynamic scoring (\cref{sec:attention_heads}), allowing for an adaptive reweighting of visual and textual features.

\subsection{Interaction signature generation}
\label{sec:prompts_expansion}

Given some visual $\mathbf{x}^{v}$ and textual $\mathbf{x}^{t}$ interaction information, the visual $\phi^{v}$ and textual $\phi^{t}$ encoders that comprise CLIP-like~\cite{Radford2021} VLM models enable their projection to a shared representation space as $\mathbf{z}^{v} = \phi^{v}(\mathbf{x}^{v})$ and $\mathbf{z}^{t} = \phi^{t}(\mathbf{x}^{t})$, where they are comparable. 
However, as explored in previous works~\cite{Momeni2023,Yuksekgonul2023,Bhalla2024}, encoders $\phi^{v}$ and $\phi^{t}$ fail to adequately capture certain textual and visual concepts. 
This limitation arises primarily from the CLIP training process, which emphasizes objects and nouns while neglecting factors such as camera orientation and distinctions between synonyms.
Therefore, we propose a method for constructing a more action-centric representation.

In the context of language modeling, the linear representation hypothesis posits that many semantic concepts can be approximated as linear functions of model representations~\cite{Mikolov2013,Park2024}, allowing the definition of mapping functions $\zeta^{t}$ and $\zeta^{v}$ that generate textual and visual information given a series of concepts.
Following this framework, the contents of a text can be expressed as $\mathbf{x}^{t} = {\zeta}^{t}(\omega, \epsilon)$, where ${\omega}$ represents the semantic concepts (\eg, animals, plants, and objects) and $\epsilon$ represents the non-semantic ones (\eg, lighting conditions, styles, and movement), as in~\cite{Bhalla2024}.
Given that CLIP is optimized to satisfy the alignment condition $\forall ~ \epsilon, \epsilon', \ \ \phi^{v}(\zeta^{v}(\omega, \epsilon)) = \phi^{t} (\zeta^{t}(\omega, \epsilon'))$, we can reasonably infer that CLIP captures semantic concepts $\omega$ while remaining invariant to $\epsilon$:
\begin{equation}
\phi^{t}(\zeta^{t}(\omega)) \approx \phi^{t}(\zeta^{t}(\omega, \epsilon)).
\end{equation}
This observation provides insight into the difficulty VLMs encounter in comprehending verbs.
Since verbs typically convey dynamic or relational information rather than static semantic content, their representations are inherently weaker compared to those of nouns and adjectives~\cite{Momeni2023}.

Following the linear representation hypothesis, we can further decompose $\epsilon = (\sigma, \epsilon^\ast)$, where $\sigma$ is the action information and $\epsilon^\ast$ are the remaining non-semantic concepts.
Although interpreting $\sigma$ proves to be ill-posed for CLIP-like encoders, it still contains relevant information that can be understood by humans and LLMs alike~\cite{Petersen2023}.
We exploit this fact and construct a set $\mathcal{T} = \{\tau_i\}_{i=1}^M$ of parameterized templates for the extraction of semantic information~\cite{Luo2024,Abdelhamed2024} and combine them with $\sigma$ through a substitution morphism~\cite{Pratt2023}:
\begin{equation}
\Theta: \mathcal{T} \times \sigma \to \mathcal{T}_\sigma, \ \ \ \Theta(\tau_i, \sigma) = \tau_i \circ \zeta^{t}(\sigma)
\end{equation}
yielding a set $\mathcal{T}_\sigma$ of completed templates.

At this point, we can leverage an LLM $\psi$ to process $\mathcal{T}_\sigma$ and generate descriptions $\mathbf{x}^{\sigma} = \psi(\mathcal{T}_\sigma)$, whose action concepts will be grounded in higly-semantic tokens.
Subsequently, we can decompose $\mathbf{x}^{\sigma} = {\zeta}^{t}(\omega^\sigma, \epsilon^\sigma)$ to isolate these action-semantic tokens $\omega^\sigma$ and combine them with the original object-related concepts $\omega$, creating new semantically-rich interaction descriptions $\hat{\mathbf{x}}^{t} = {\zeta}^{t}(\omega, \omega^\sigma)$.

\begin{figure}[!t]
\centering
\includegraphics[width=0.8\linewidth]{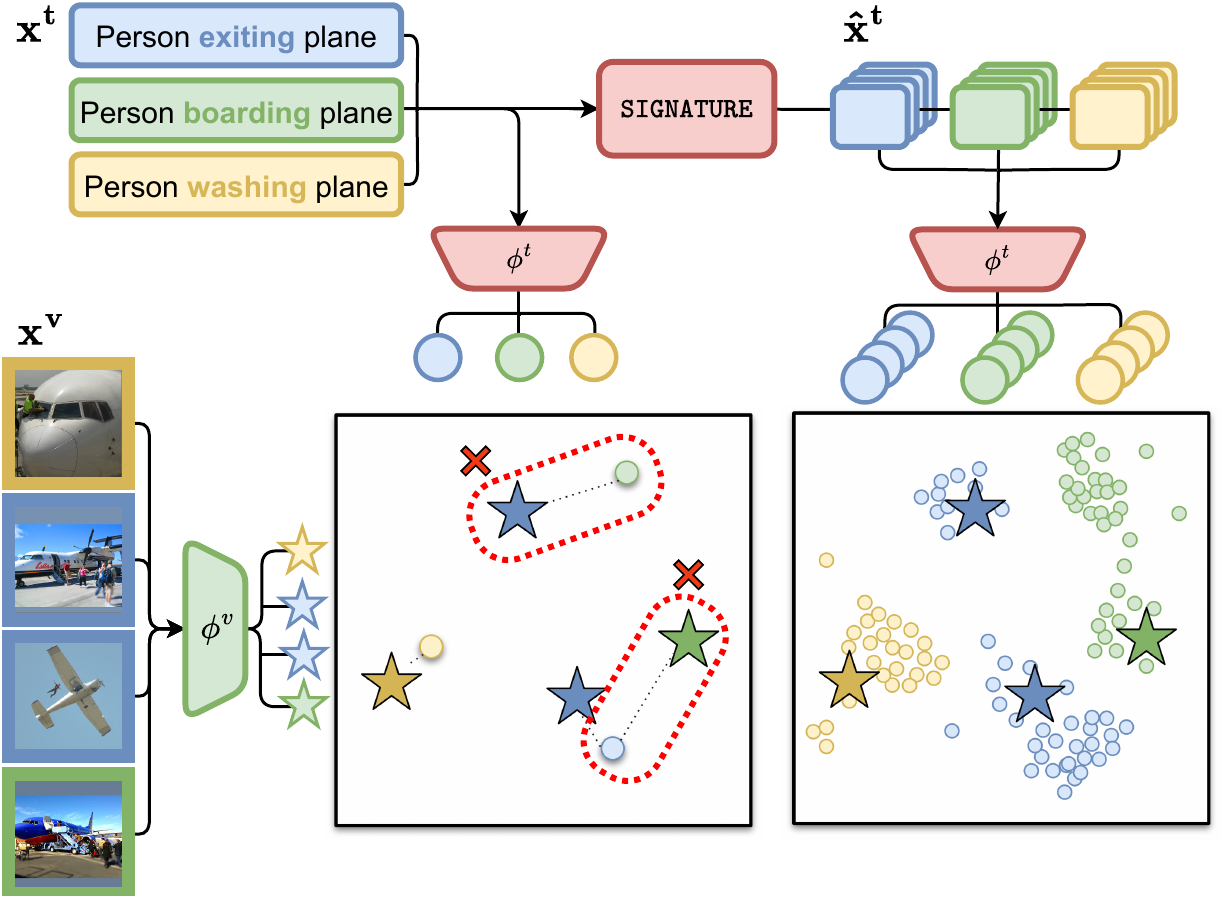}
\caption{Using standard HOI textual information for the prediction of interactions often leads to incorrect associations (left).
To remedy this, the \MyColorBox[ourred]{signature generation} process of \method extracts highly-semantic information from interactions (right).
This results in a distribution in the CLIP embedding space that is more aligned with the visual features and enables the representation of complex stratified manifolds instead of being limited to a single point.
Different colors represent different interactions.
\Description{The signature generation process of \method extracts highly-semantic information from interactions.}
\vspace{-1.5em}
}
\label{fig:signature}
\end{figure}

This new information can now be projected to the shared representation space $\hat{\mathbf{z}}^{t} = \phi(\hat{\mathbf{x}}^{t}) \in \mathbb{R}^{M \times d}$, creating the \textbf{interaction signature} of $\mathbf{x}^{t}$, which no longer will be independent of the verbal information.
$\hat{\mathbf{z}}^{t}$ can be precomputed once per interaction and reused multiple times and, differently from other representation techniques~\cite{Koh2020,Maiorca2023}, does not rely on either the training set or reduces our signature to one single vector, allowing a more flexible representation of complex stratified manifolds, as shown in \cref{fig:signature}.

\subsection{Human-object pair generation}
\label{sec:detector}
Given an RGB image $\mathcal{I}$, the first stage of \method aims to detect all potential human-object pairs.
To this end, we employ a frozen DETR~\cite{Carion2020} detector, following standard practice in recent HOI detection~\cite{Lei2023,Zhang2022,Mao2023,Ning2023}.
Formally, we obtain a set $\mathcal{O} = \{o_i\}_{i=1}^N$ of $N$ detections, where each detection $o_i = (c_{x},~c_{y},~h,~w,~l)$ is defined by its center coordinates $(c_{x},~c_{y})$, height $h$, width $w$, and label $l$.
Subsequently, we obtain a set of human detections $\mathcal{O}_H = \{ o \in \mathcal{O} \mid l = \text{\texttt{"human"}} \}$ and construct all the possible interactions pairs $\mathcal{P} = \{ \langle o_h,~o_o\rangle ~ \forall ~ o_h \in \mathcal{O}_H, ~ o_o \in \mathcal{O} \mid o_h \neq o_o \}$.

Let $\Psi(\mathcal{I}, o)$ represent the operation that crops the portion of image $\mathcal{I}$ corresponding to the bounding box delimited by detection $o$.
Given a pair $\langle o_h,~o_o \rangle$, we can extract the visual features of the human $\mathbf{z}^{h} \in \mathbb{R}^{d}$, the object $\mathbf{z}^{o} \in \mathbb{R}^{d}$, and their union $\mathbf{z}^{u} \in \mathbb{R}^{d}$ as:
\begin{align}
\mathbf{z}^{h} & = \phi^{v}(\Psi(\mathcal{I}, o_h)) \label{eq:test-human-features}\\
\mathbf{z}^{o} & = \phi^{v}(\Psi(\mathcal{I}, o_o)) \label{eq:test-object-features}\\
\mathbf{z}^{u} & = \phi^{v}(\Psi(\mathcal{I}, o_h \cup o_o)) \label{eq:test-human_object-features}
\end{align}
In its second stage, \method will process each element in $\mathcal{P}$ and leverage their visual features to assign it an interaction class.

\subsection{Multi-head attention}
\label{sec:attention_heads}

Inspired by Transformer architectures~\cite{vaswani2017attention}, where a single attention head may not suffice to capture complex relationships, we recast HOI prediction as a multi-head attention process.
In our formulation, the multi-head mechanism processes the visual features of a human-object pair, with each head specializing in extracting distinct aspects.
Textual heads rely on the generated interaction signatures $\mathbb{S}$ to focus on capturing both \textit{fine}-grained and \textit{coarse} semantics about the interaction.
Visual heads, on the other hand, analyze the visual appearance of the involved \textit{instances} and their \textit{contextual} environment.
Additionally, we introduce a negative bias to visual features to better distinguish visually similar interactions.
This design enables \method to flexibly prioritize fine-grained details or broader contextual cues according to the interaction scenario. \\

\vspace{-0.5em}
\noindent \textbf{Attention design.}
The attention heads adopted by \method are designed to closely resemble the structure of the standard attention mechanism introduced by~\cite{vaswani2017attention}.
Specifically, given a query matrix $\mathbf{q}_h \in \mathbb{R}^{1 \times d_h}$, a key matrix $\mathbf{k}_h \in \mathbb{R}^{s_h \times d_h}$, and a value matrix $\mathbf{v}_h \in \mathbb{R}^{s_h \times I}$, the attention output $\mathbf{a}_h \in \mathbb{R}^{I}$ for the $h$-th head is computed as:
\begin{equation}
    \mathbf{a}_h = \left( \mathbf{q}_h  \mathbf{k}^T_h \right) \mathbf{v}_h,
\end{equation}
where $I$ denotes the number of interaction classes, $d_h$ is the feature dimension, and $s_h$ is the number of classification samples.
In our setup, $\mathbf{q}_h$ corresponds to the human-object pair requiring classification, $\mathbf{k}_h$ contains the visual or textual sample features used for classification, and $\mathbf{v}_h$ consists of one-hot encoded interaction labels for each sample in $\mathbf{k}_h$. \\

\vspace{-0.5em}
\noindent \textbf{\method's multi-head configuration.}
Leveraging distinct input matrices for each head, \method extracts complementary perspectives that enhance interaction prediction. 
We find that constructing our multi-head predictor with the following configuration yields an optimal balance between simplicity and performance:

\noindent \textbf{Textual fine-grained head ($H^T_F$):} 
This inter-modal head focuses on subtle semantic similarities.
It employs text-based interaction signatures (\cref{sec:prompts_expansion}) as keys $\mathbf{k} = \hat{\mathbf{z}}^{t}$ and human-object union features (\cref{eq:test-human_object-features}) as queries $\mathbf{q} = \mathbf{z}^{u}$.
Accordingly, the feature dimension of the head is the same as the shared representation space $d_h = d$, whereas $s_h = M$ is the number of parametrized templates.

\noindent \textbf{Textual coarse head ($H^T_C$):}
This inter-modal head provides a broader, more general interaction perspective.
It uses the averaged interaction signatures as keys $\mathbf{k_i} = \frac{1}{M} \sum_{j=1}^{M} \hat{\mathbf{z}}^{t}_{j, i}$, with the same human-object union features as queries $\mathbf{q} = \mathbf{z}^{u}$.
Thus, the dimensions of this head are $d_h = d$ and $s_h = 1$.

\noindent \textbf{Visual instance head ($H^V_I$):}
This intra-modal head captures fine-grained details by focusing on the human and the object independently, rather than on their union.
It utilizes a small registry of HOI interaction samples $\mathcal{R} = \{\langle \tilde{o}_{h_i},~\tilde{o}_{o_i}\rangle\}_{i=1}^J$, with $\tilde{o_h}, \tilde{o_o}$ corresponding to humans and objects from images of the small registry, keys generated by concatenating their visual features $\mathbf{k} = \phi^{v}(\tilde{o}_{h}) \mathbin\Vert \phi^{v}(\tilde{o}_{o})$, and queries formed by concatenating image features of human and object instances (\cref{eq:test-human-features,eq:test-object-features}) as queries $\mathbf{q} = \mathbf{z}^{h} \mathbin\Vert \mathbf{z}^{o}$.
Consequently, the feature dimension of the head becomes $d_h = 2d$, and the number of samples is $s_h = J$.

\noindent \textbf{Visual contextual head ($H^V_C$):}
To capture the broader contextual environment where interactions take place, this intra-modal head uses the union of human-object bounding box visual features as keys $\mathbf{k}~=~\phi^{v}(\tilde{o}_{h} \cup \tilde{o}_{o})$, and the union image features as queries $\mathbf{q} = \mathbf{z}^{u}$.
Therefore, $d_h = d$ and $s_h = 1$ for this head.

\noindent \textbf{Negative bias ($\mathbb{N}$):}
To improve performance on interactions that are
visually similar (\eg \texttt{``eating broccoli''} vs \texttt{``smelling broccoli''}) (see Supp. Mat.),
we introduce a negative attention bias.
For each visual attention head, this bias is computed as $\mathbb{N}_h = - ( \mathbf{q}_h  \mathbf{k}^T_h ) ( 1 - \mathbf{v}_h )$ and is added to the attention output to enhance the contrast between interactions.

This configuration enables each head to specialize, contributing either fine-grained or coarse-grained cues from textual and visual data.
This modularity substantially enhances \method's capacity to robustly interpret and predict a wide range of interactions. \\

\vspace{-0.5em}
\noindent \textbf{\multiheadacro (\multihead).}
Unlike previous works that aggregate multiple information streams directly~\cite{Chao2018,Gao2020}, \method leverages a \multiheadacro (\multihead), which dynamically adjusts each head's contribution to the final prediction.
Given our set of $N=4$ attention heads $\mathcal{H} = \{ H^T_F, H^T_C, H^V_I, H^V_C \}$, \multihead computes a contribution matrix $C \in \mathbb{R}^{N \times I}$, where each element $C_{h,i}$ represents the importance of head $h$ for interaction class $i$.
This is defined as
\begin{equation}
    C_{h,i} = \frac{e^{\frac{a_{h,i}}{\tau}}}{1 + \sum_{k=1}^{N} e^{\frac{a_{k,i}}{\tau}}},
\end{equation}
where $\tau$ is a temperature parameter that modulates the sharpness of the resulting distribution, ensuring that the most relevant heads contribute more significantly.
The final interaction probabilities $\mathbf{p} \in \mathbb{R}^{I}$ are then computed by weighting the heads' outputs using the contribution matrix.
Thus, given the attention outputs $A = (\mathbf{a}_{h_1}^T, \mathbf{a}_{h_2}^T, \dots, \mathbf{a}_{h_N}^T)^T \in \mathbb{R}^{N \times I}$, the probabilities are computed as
\begin{equation}
    \mathbf{p} = \frac{1}{N}\left( A \ \odot \ (1+C) \right)^{T},
\end{equation}
where $\odot$ denotes the Hadamard product.
By weighting the outputs in this manner, \multihead selectively amplifies the contributions of the most relevant heads based on interaction-specific cues.

\begin{table*}[!ht]
\centering
\caption{\textbf{SOTA comparison on HICO-DET~\cite{Chao2018} and V-COCO~\cite{Gupta2015}.} Original ADA-CM~\cite{Lei2023} results are in \textcolor{badgray}{gray}. $\dagger$ denotes results recomputed using the official code. The top-performing training-free methods are marked in \textbf{bold}, while the best training-based methods are \underline{underlined}. Note that only the latest results for training-based methods are reported here; for an extended list, please refer to \cite{Wu2024b}.}
\resizebox{0.82\textwidth}{!}{%
\begin{tabular}{@{}l@{\hskip 30pt}c@{\hskip 24pt}c@{\hskip 24pt}c@{\hskip 24pt}c@{\hskip 10pt}c@{\hskip 24pt}c@{\hskip 24pt}c@{}}
\toprule
\multicolumn{1}{c}{\multirow{2}{*}{\textbf{Method}}} & \multicolumn{4}{c}{\textbf{HICO-DET}~\cite{Chao2018}}                                           & \multicolumn{1}{l}{\textbf{}} & \multicolumn{2}{c}{\textbf{V-COCO}~\cite{Gupta2015}} \\ \cmidrule(l){2-8} 
\multicolumn{1}{c}{}                        & \textbf{Rare}                & \textbf{Non-rare}            & \textbf{AFull}               & \textbf{Full}                & \multicolumn{1}{l}{\textbf{}} & $\bm{AP^{S1}_{role}}$     & $\bm{AP^{S2}_{role}}$     \\ \midrule
\multicolumn{8}{c}{\cellcolor[HTML]{a9d6e5}Training-based - One stage}                                                                                                                                            \\ \midrule
Iwin~\cite{Tu2022}                          & 27.62\%             & 34.14\%             & 30.88\%             & 32.03\%             &                               & 60.47\%              & --                   \\
CDN~\cite{zhang2021mining}                  & 27.19\%             & 33.53\%             & 30.36\%             & 32.07\%             &                               & 61.68\%              & 63.77\%              \\
GEN-VLKT~\cite{Liao2022}                    & 29.25\%             & 35.10\%             & 32.17\%             & 33.75\%             &                               & 62.41\%              & 64.46\%              \\ \midrule
\multicolumn{8}{c}{\cellcolor[HTML]{a9d6e5}Training-based - Two stage}                                                                                                                                            \\ \midrule
HOICLIP~\cite{Ning2023}                     & 31.12\%             & 35.74\%             & 33.43\%             & 34.69\%             &                               & 63.50\%              & 64.80\%              \\
CLIP4HOI~\cite{Mao2023}                     & 33.95\%             & 35.74\%             & 34.84\%             & 35.33\%             &                               & --                   & 66.30\%              \\
SICHOI~\cite{Luo2024}            & \underline{42.38}\% & \underline{41.61}\% & \underline{41.99}\% & \underline{41.79}\% &                               & \underline{67.90}\%  & \underline{72.80}\%  \\
BCOM~\cite{wang2024bilateral} & 39.90\% & 39.17\% & 39.54\% & 39.34\% & & 65.80\% & 69.90\% \\
Wu et al.~\cite{Wu2024b}                    & 32.48\%             & 36.86\%             & 34.67\%             & 35.86\%             &                               & 61.10\%              & 66.60\%              \\ \midrule
\multicolumn{8}{c}{\cellcolor[HTML]{a9d6e5}Training-free}                                                                                                                                                         \\ \midrule
CLIP ViT-B/16~\cite{Radford2021}            & 27.79\% & 	19.25\% & 	23.52\% & 	21.21\% & & 		35.83\% & 	40.63\%              \\
CLIP ViT-L/14~\cite{Radford2021}            & 30.97\%	& 19.65\%	& 25.31\%	& 22.26\%		& & 38.44\%	& 43.45\%              \\
LongCLIP-B~\cite{Zhang2024}                 & 28.27\%	& 20.13\%	& 24.20\%	& 22.00\%		& & 36.26\%	& 41.00\%              \\
LongCLIP-L~\cite{Zhang2024}                 & 31.32\%	& 20.68\%	& 26.00\%	& 23.13\%		& & 40.13\%	& 45.11\%              \\
\textcolor{badgray}{ADA-CM~\cite{Lei2023}}                       & \textcolor{badgray}{27.24\%}             & \textcolor{badgray}{24.58\%}             & \textcolor{badgray}{25.91\%}             & \textcolor{badgray}{25.19\%}             &                               & \textcolor{badgray}{39.09\%}              & \textcolor{badgray}{43.93\%}              \\
ADA-CM$\dagger$~\cite{Lei2023}              & 27.61\%             & 24.48\%             & 26.04\%             & 25.20\%             &                               & 38.68\%              & 43.51\%              \\[-1pt]

\midrule

\textbf{\method (Ours)}                     & \textbf{34.22\%}    & \textbf{26.46\%}    & \textbf{30.34\%}    & \textbf{28.24\%}    &                               & \textbf{42.80\%}     & \textbf{47.82\%}     \\ \bottomrule
\end{tabular}%
}
\label{tab:main_results}
\end{table*}

\section{Experiments} \label{sec:experiments}

\subsection{Experimental setting}
\label{sec:exp}
\noindent \textbf{Datasets.}
Our experiments are carried on the V-COCO~\cite{Gupta2015} and HICO-DET~\cite{Chao2018} datasets.
V-COCO, which is a subset of COCO, comprises 10,396 images, split into 5,400 train-val images and 4,946 test images depicting 24 action types and 80 object classes \cite{hou2021detecting,Lei2023}. 
HICO-DET contains 47,776 images, 38,118 for training and 9,658 for testing. It includes 117 action types and 80 object classes, for a total of 600 HOI categories. \\

\noindent \textbf{Evaluation Metrics.}
In line with the standard practice, we measure model performance using the mean average precision (mAP).
For V-COCO, we report the average precision (AP) under two conditions: $AP_{role}^{S1}$, which evaluates all actions regardless of whether they involve an object (e.g., \texttt{``hold a cup''}, \texttt{``stand''}, or \texttt{``smile''}), and $AP_{role}^{S2}$, which considers only interactions where the action involves a specific object (e.g., \texttt{``cut with a knife''} or \texttt{``sit on a chair''}).
We report the mAP for the HICO-DET dataset across its two main categories: 138 HOI categories with fewer than 10 training samples (Rare) and 462 HOI categories (Non-rare). 
Furthermore, consistent with training-free HOI literature~\cite{Lei2023}, we evaluate our model’s zero-shot performance on the HICO-DET dataset using two settings: (1) Rare first setting (RF)~\cite{hou2021detecting}, which prioritizes rare HOI categories when selecting held-out triplets, and (2) Non-rare first setting (NF)~\cite{hou2021detecting}, which prioritizes non-rare HOI categories, resulting in a smaller, more challenging test set. 
We provide both the weighted average (Full) and the arithmetic average (AFull) across all 600 HOI categories.

\subsection{Implementation details}
In line with established practices in HOI detection~\cite{Lei2023,Zhang2022,Mao2023,Ning2023}, \method employs a 
frozen object detector to identify all humans and objects within the scene.
We filter detections with a confidence threshold below $0.2$ and sample a minimum of 3 and a maximum of 15 human and object instances.
Unless otherwise stated, we utilize CLIP~\cite{Radford2021} encoders as the vision ($\phi^v$) and textual ($\phi^t$) backbones, while \cite{OpenAI2023} is leveraged as $\psi$ during the interaction signature generation.
For a fair comparison, we adopt the same hyperparameter settings and backbone configurations as those used in ADA-CM~\cite{Lei2023}.
We set the temperature of the \multihead module to $\tau = 0.1$ and the maximum size of the registry for each interaction to $J = 8$.
To generate our interaction signatures, as detailed in \cref{sec:prompts_expansion}, each signature is represented as a matrix of dimensions $M \times d$, where $d$ corresponds to the dimensionality of the shared representation space of $\phi^{v}$ and $\phi^{t}$, while $M = 50$ is empirically determined as described in \cref{sec:ablation}.
Refer to \cref{sec:pseudolabels} for more details on how the registry can be generated, and Supp. Mat. for extra implementation details and visualizations of our generated interaction signatures.

\begin{table}[!t]
\centering
\caption{\textbf{Zero-shot experiments on HICO-DET~\cite{Chao2018}} with CLIP ViT-B/16. RF = rare first. NF = non-rare first. Original ADA-CM~\cite{Lei2023} results are in \textcolor{badgray}{gray}. $\dagger$ denotes results recomputed using the official code. Best results are in \textbf{bold}.}
\resizebox{0.9\columnwidth}{!}{%
\begin{tabular}{@{}l@{\hskip 4pt}c@{\hskip 7pt}cccc@{}}
\toprule
\textbf{Method}     & \textbf{Setting} & \textbf{Seen}    & \textbf{Unseen}  & \textbf{AFull}        & \textbf{Full}    \\ \midrule
ADA-CM~\cite{Lei2023}          & \textcolor{badgray}{RF}      & \textcolor{badgray}{24.54\%} & \textcolor{badgray}{26.83\%} & \textcolor{badgray}{25.68\%}          & \textcolor{badgray}{25.00\%} \\
ADA-CM$\dagger$~\cite{Lei2023} & RF      & 21.55\% & 26.73\% & 24.14\%          & 22.59\% \\
\textbf{\method (Ours)}                 & RF      & \textbf{23.96\%}            & \textbf{30.36\%}            & \textbf{27.16\%} & \textbf{25.24\%}            \\ 
\midrule
\textcolor{badgray}{ADA-CM~\cite{Lei2023}}          & \textcolor{badgray}{NF}      & \textcolor{badgray}{23.16\%}                     & \textcolor{badgray}{30.11\%}            & \textcolor{badgray}{26.63\%} & \textcolor{badgray}{24.55\%}                     \\
ADA-CM$\dagger$~\cite{Lei2023} & NF      & 23.79\%                     & 26.13\%                     & 24.96\%          & 24.26\%                     \\
\textbf{\method (Ours)}                 & NF      & \textbf{24.58\%}            & \textbf{27.56\%}                     & \textbf{26.07\%}          & \textbf{25.18\%}    \\ \bottomrule
\end{tabular}%
}
\label{tab:hicodet_rf_nf}
\vspace{-1.5em}
\end{table}

\subsection{Comparison with the state-of-the-art}
\label{sec:SOTAcomp}
Currently, ADA-CM~\cite{Lei2023} stands as the only training-free method for HOI detection.
Nonetheless, since both ADA-CM and our approach leverage VLMs, it is logical also to evaluate the HOI performance of related VLM-based methods, such as CLIP~\cite{Radford2021} and LongCLIP~\cite{Zhang2024}.
We present the results for both datasets in \cref{tab:main_results}, alongside the top-performing training-based methods from the current state of the art.
Among the training-free approaches, our proposal, \method, achieves the highest performance across all metrics.
Remarkably, for the Rare class setting, \method surpasses all training-based methods but ~\cite{Luo2024,wang2024bilateral}, demonstrating its strong performance in challenging categories.
In \cref{tab:hicodet_rf_nf}, we compare the performance of our \method against that of ADA-CM in the zero-shot setting, as described in \cref{sec:exp}.
Our \method consistently outperforms ADA-CM~\cite{Lei2023} across all conditions.
Notably, our model achieves higher performance on unseen categories compared to seen categories, demonstrating that our approach successfully addresses the lack of visual features when predicting unseen interactions.
This improvement is particularly pronounced in the rare first (RF) setting, further underscoring the robustness of \method in zero-shot settings.

\subsection{Ablation study}
\label{sec:ablation}

\begin{table}[!t]
\centering
\caption{\textbf{Ablations on HICO-DET} with CLIP ViT-B/16 \cite{Radford2021} as the VLM backbone. $H^V_C$ is the visual contextual head, $H^V_I$ is the visual instance head, $\mathbb{N}$ is the negative bias, $H^T_C$ is the textual coarse head, $H^T_F$ is the textual fine-grained head.}
\resizebox{0.9\columnwidth}{!}{%
\begin{tabular}{@{}c@{\hskip 4pt}c@{\hskip 4pt}c@{\hskip 4pt}c@{\hskip 4pt}c@{\hskip 2pt}c@{\hskip 2pt}c@{\hskip 2pt}c@{\hskip 2pt}c@{\hskip 4pt}c@{}}
\toprule
$\bm{H^V_C}$ & $\bm{H^V_I}$ & $\bm{\mathbb{N}}$ & $\bm{H^T_C}$ & $\bm{H^T_F}$ & \textbf{\multihead} & \textbf{Rare}             & \textbf{Non-rare}         & \textbf{AFull}            & \textbf{Full}             \\ \midrule
\xmark  & \cmark & \cmark  & \cmark  & \cmark       &\cmark     & \textbf{30.59\%}	& 24.72\%	& 27.65\%	& 26.07\%          \\
\cmark  & \xmark & \cmark  & \cmark  & \cmark         &\cmark     & 28.49\%	& 24.49\%	& 26.49\%	& 25.41\%          \\
\cmark  & \cmark & \xmark  & \cmark  & \cmark         &\cmark     & 29.91\%	& 24.84\%	& 27.38\%	& 26.01\%         \\
\cmark  & \cmark & \cmark  & \xmark  & \cmark        & \cmark     & 29.13\%	& 24.47\%	& 26.80\%	& 25.54\%          \\
\cmark  & \cmark & \cmark  & \cmark  & \xmark        & \cmark     & 28.89\%	& 24.44\%	& 26.66\%	& 25.46\%          \\
\cmark  & \cmark & \cmark  & \cmark  & \cmark        & \xmark     & 29.31\%	& 24.89\%	& 27.13\%	& 25.75\%          \\
\cmark       & \cmark & \cmark  & \cmark  & \cmark  & \cmark     & 30.53\%          & \textbf{24.92\%} & \textbf{27.73\%} & \textbf{26.21\%} \\ \bottomrule
\end{tabular}%
}
\label{tab:hicodet_ablations}
\end{table}

\textbf{Multi-head attention.} \cref{tab:hicodet_ablations} illustrates the ablation study conducted on the HICO-DET dataset, analyzing the impact of each component within our \method.
The findings demonstrate that the integration of all model heads, along with the negative bias $\mathbb{N}$ and \multihead,
yields the highest overall performance, particularly improving the model's capability in Rare and Full settings.
Notably, each component independently contributes to improving the model’s overall performance, underscoring their individual and collective importance.
The largest performance decrease relative to the full configuration of \method (shown in the last row) occurs when either the visual instance head ($H_I^V$) or the textual fine head ($H^T_F$) are disabled (second and fifth row), with all other elements enabled.
In this scenario, full accuracy drops to just 25.41\% and 25.46\%, respectively.
This emphasizes the critical role that rich semantic information plays in HOI detection. \\

\vspace{-0.5em}
\noindent \textbf{Impact of $\tau$ in attention selection.}
\begin{table}[!t]
\centering
\caption{\textbf{Effect of temperature $\bm{\tau}$} on HICO-DET~\cite{Chao2018} with CLIP ViT-B/16.}
\resizebox{0.7\columnwidth}{!}{%
\begin{tabular}{@{}ccccc@{}}
\toprule
$\bm{\tau}$ & \textbf{Rare}             & \textbf{Non-rare}         & \textbf{AFull}         & \textbf{Full}             \\ \midrule
1.0    & 29.77\%	& 24.96\%	& 27.37\%	& 26.07\%          \\
0.8    & 29.92\%	& 24.95\%	& 27.44\%	& 26.10\%          \\
0.6    & 29.99\%	& 24.99\%	& 27.49\%	& 26.14\%          \\
0.4    & 30.20\%	& 24.99\%	& 27.60\%	& 26.19\%          \\
0.2    & 30.30\%	& \textbf{25.00\%}	& 27.65\%	& \textbf{26.22\%}          \\
0.1    & \textbf{30.53\%}	& 24.92\%	& \textbf{27.73\%}	& 26.21\% \\ \bottomrule
\end{tabular}%
}
\label{tab:hicodet_ablations_temp}
\end{table}

As described in \cref{sec:attention_heads}, \method's \multihead incorporates a temperature parameter $\tau$ to regulate the smoothness of the contributions of the different heads.
Results in \cref{tab:hicodet_ablations_temp} demonstrate that lower values of $\tau$ yield better performance, while increments above $0.4$ show negligible effect on non-rare and rare predictions for the HICO-DET dataset. \\

\vspace{-0.5em}
\noindent \textbf{Influence of the VLM backbone.}
\begin{table}[!t]
\centering
\caption{\textbf{Effect of different backbones on \method} evaluated on ADA-CM~\cite{Lei2023}. $\dagger$ denotes results recomputed using the official code. The best results are marked in \textbf{bold}.}
\resizebox{0.95\columnwidth}{!}{%
\begin{tabular}{@{}l@{\hskip 4pt}c@{\hskip 1pt}c@{\hskip 2pt}c@{\hskip 2pt}c@{\hskip -3pt}c@{\hskip 0pt}c@{\hskip 4pt}c@{}}
\toprule
\multicolumn{1}{c}{}                         & \multicolumn{4}{c}{\textbf{HICO-DET}~\cite{Chao2018}}                               & \multicolumn{1}{l}{} & \multicolumn{2}{c}{\textbf{V-COCO}~\cite{Gupta2015}} \\  \cmidrule(l){2-8} 
\multicolumn{1}{c}{\multirow{-2}{*}{\textbf{Method}}} & \textbf{Rare}             & \textbf{Non-rare}         & \textbf{AFull}            & \textbf{Full}             & \multicolumn{1}{l}{} & $\bm{AP^{S1}_{role}}$     & $\bm{AP^{S2}_{role}}$     \\ \midrule
\multicolumn{8}{c}{\cellcolor[HTML]{A9D6E5}ViT-B/16}                                                                                                                                          \\ \midrule
ADA-CM$\dagger$~\cite{Lei2023}               & 27.61\%          & 24.48\%          & 26.04\%          & 25.20\%          &                      & 38.68\%              & 43.51\%              \\
\textbf{\method (Ours)}                               & 30.53\%          & 24.92\%          & 27.73\%          & 26.21\%          &                      & 40.14\%              & 45.00\%              \\ 
\midrule
\multicolumn{8}{c}{\cellcolor[HTML]{A9D6E5}LongCLIP-B}                                                                                                                                        \\ \midrule
ADA-CM$\dagger$~\cite{Lei2023}               & 27.94\%          & 25.08\%          & 26.51\%          & 25.73\%          &                      & 39.46\%              & 44.25\%              \\
\textbf{\method (Ours)}                               & 29.45\%          & 25.52\%          & 27.48\%          & 26.42\%          &                      & 40.04\%              & 44.82\%               \\ 
\midrule
\multicolumn{8}{c}{\cellcolor[HTML]{A9D6E5}ViT-L/14}                                                                                                                                          \\ \midrule
ADA-CM$\dagger$~\cite{Lei2023}               & 31.54\%          & 26.01\%          & 28.78\%          & 27.28\%          &                      & 40.11\%              & 44.91\%              \\
\textbf{\method (Ours)}                               & \textbf{34.22\%} & 26.46\%          & 30.34\% & 28.24\%          &                      & 41.02\%              & 45.80\%              \\ 
\midrule
\multicolumn{8}{c}{\cellcolor[HTML]{A9D6E5}LongCLIP-L}                                                                                                                                        \\ \midrule
ADA-CM$\dagger$~\cite{Lei2023}               & 31.49\%          & 27.36\%          & 29.42\%          & 28.31\%          &                      & 42.51\%              & 47.47\%               \\

\textbf{\method (Ours)}                               & 33.63\%          & \textbf{27.62\%} & \textbf{30.63\%}          & \textbf{29.00\%} &                      & \textbf{42.80\%}     & \textbf{47.82\%}     \\ 
\bottomrule
\end{tabular}%
}
\label{tab:hicodet_abl_backbones}
\vspace{-2em}
\end{table}

We further investigate the effect of four different VLM backbones on the performance of both ADA-CM and our proposed method, as shown in \cref{tab:hicodet_abl_backbones}.
Our approach consistently achieves superior results across various architectures, including ViT-B/16 and ViT-L/14, which were trained using different pretraining strategies~\cite{Radford2021,Zhang2024}. Notably, as the backbone architecture grows in size and complexity, our method demonstrates an enhanced ability to extract meaningful features, leading to systematic improvements over the current SOTA. \\

\vspace{-0.5em}
\noindent \textbf{Effect of our interaction signatures.}
\begin{table}[!t]
\centering
\caption{\textbf{Effect of injecting our interaction signatures $\bm{\mathbb{S}}$ into different VLM backbones} on the HICO-DET~\cite{Chao2018} dataset.}
\resizebox{0.9\columnwidth}{!}{%
\begin{tabular}{@{}lcccc@{}}
\toprule
\multicolumn{1}{c}{\textbf{Method}}                              & \textbf{Rare}             & \textbf{Non-rare}         & \textbf{AFull}            & \textbf{Full}             \\ \midrule
CLIP ViT-B/16~\cite{Radford2021}     & 27.13\%	& 19.25\%	& 24.09\%	& 21.21\%          \\
\plusours{} \textbf{(Ours)}     & \textbf{28.14\%}	& \textbf{20.70\%}	& \textbf{24.42\%}	& \textbf{22.41\%}          \\[5pt]

CLIP ViT-L/14~\cite{Radford2021}     & 30.97\%	& 19.65\%	& 25.31\%	& 22.26\%          \\
\plusours{} \textbf{(Ours)}     & \textbf{31.23\%}	& \textbf{20.89\%}	& \textbf{26.06\%}	& \textbf{23.27\%}          \\[5pt]

LongCLIP-B~\cite{Zhang2024}         & 28.27\%	& 20.13\%	& 24.20\%	& 22.00\%          \\
\plusours{} \textbf{(Ours)}          & \textbf{29.43\%}	& \textbf{22.33\%}	& \textbf{25.88\%}	& \textbf{23.96\%}          \\[5pt]

LongCLIP-L~\cite{Zhang2024} & 31.32\%	& 20.68\%	& 26.00\%	& 23.13\%           \\[-1pt]
\plusours{} \textbf{(Ours)} & \textbf{33.08\%}	& \textbf{23.24\%}	& \textbf{28.16\%}	& \textbf{25.50\%}          \\[2pt]

\midrule

\textbf{\method (Ours)}                                 & \textbf{34.22\%}    & \textbf{26.46\%}    & \textbf{30.34\%}    & \textbf{28.24\%} \\ \bottomrule
\end{tabular}%
}
\label{tab:hicodet_alb_signatures}
\vspace{-1em}
\end{table}

We evaluate the effect of injecting our interaction signatures into different VLM backbones.
As shown in Tab.~\ref{tab:hicodet_alb_signatures}, our signatures not only enhance our method's performance but also improve the prediction effectiveness of both CLIP~\cite{Radford2021} and LongCLIP~\cite{Zhang2024} across their different architectures.
Nevertheless, all these methods still fall short of matching the performance of our \method. \\

\noindent \textbf{Impact of the signature dimensionality.}
We analyze the impact of interaction signature dimensionality in \cref{tab:supp_abl_singatures_per_template}, evaluating different values of $M$, specifically $M \in \{5, 10, 25, 50\}$.
This experiment is conducted on the HICO-DET dataset~\cite{Gupta2015} using ViT-B/16 as the backbone.
Our results indicate that increasing the dimensionality of the signatures consistently improves performance for both rare and non-rare interactions.
This trend is intuitive, as higher dimensionality allows for richer semantic representations and greater flexibility in capturing the interaction manifolds. 
\begin{table}[!ht]
\centering
\vspace{-0.5em}
\caption{\textbf{Effect of the dimensionality of signatures $\bm{\mathbb{S}}$} on \method.
Tested on HICO-DET~\cite{Gupta2015} using ViT-B/16 as backbone.
The best results are marked in \textbf{bold}.
$\star$ denotes our default configuration.}
\resizebox{0.7\columnwidth}{!}{%
\begin{tabular}{lcccc}
\toprule
\textbf{$\bm{\mathbb{S}}$} & \textbf{Rare} & \textbf{Non-rare} & \textbf{AFull} & \textbf{Full} \\ \midrule
5                                                 & 30.10\%	& 24.81\%	& 27.45\%	& 26.03\%       \\
10                                                 & 30.33\%	& 24.80\%	& 24.80\%	& 26.07\%       \\
25                                                 & 30.42\%	& 24.87\%	& 27.65\%	& 26.15\%       \\
$50^{\star}$                                                & \textbf{30.53\%}       & \textbf{24.92\%}           & \textbf{27.73\%}        & \textbf{26.21\%}       \\ \bottomrule
\end{tabular}%
}
\label{tab:supp_abl_singatures_per_template}
\vspace{-1em}
\end{table}

\subsection{Label-free HOI}\label{sec:pseudolabels}
As described above, \method relies on a small registry of HOI interaction samples $\mathcal{R}$ for its visual heads $H^V_I$ and $H^V_C$.
This reliance on annotated data can also be found in training-based~\cite{Ning2023,Mao2023,Luo2024} and training-free~\cite{Lei2023} methods.
Here, we show how we can remove this assumption by introducing \methodaf, a variation of \method where no manually curated interaction labels are given.
\begin{table}[!t]
\caption{Performace results of our \methodaf. LF stands for ``label-free''. The best results for each setting are marked in bold.}
\resizebox{0.9\columnwidth}{!}{%
\begin{tabular}{@{}lccccc@{}}
\toprule
\multicolumn{1}{c}{\textbf{Method}} & \textbf{LF} & \textbf{Rare}   & \textbf{Non-rare} & \textbf{AFull}  & \textbf{Full}   \\ 
\midrule
ADA-CM$\dagger$~\cite{Lei2023} & \xmark & 27.61\% & 24.48\% & 26.04\% & 25.20\% \\
\method (Ours) & \xmark & \textbf{30.53\%} & \textbf{24.92\%}   & \textbf{27.73\%} & \textbf{25.75\%} \\
\midrule
CLIP ViT-B/16~\cite{Radford2021} & \cmark  & 27.79\% & 19.25\%   & 23.52\% & 21.21\% \\
\methodaf (Ours) & \cmark & \textbf{29.29\%} & \textbf{22.92\%}   & \textbf{26.10\%} & \textbf{24.38\%} \\
\bottomrule
\end{tabular}%
}
\label{tab:abl_pseudolabeling}
\end{table}

Specifically, \methodaf leverages its textual heads $H^T_F$ and $H^T_C$ to compute similarity scores between human-object visual pairs $\mathcal{P}$, generating interaction pseudolabel scores. We then keep only predictions exceeding a confidence threshold of $\mathbf{p} \geq 0.9$, and use those to populate the value matrix $\mathbf{v}_h$ and visual head registry. As demonstrated in \cref{tab:abl_pseudolabeling}, \methodaf achieves performance competitive with state-of-the-art methods while eliminating the need for manually annotated interaction labels.
Refer to the Supp. Mat. for additional experiments on label-free HOI.

\subsection{Qualitative results}
\label{sec:qualitatives}

\begin{figure}[!ht]
\centering
\includegraphics[width=1.0\linewidth]{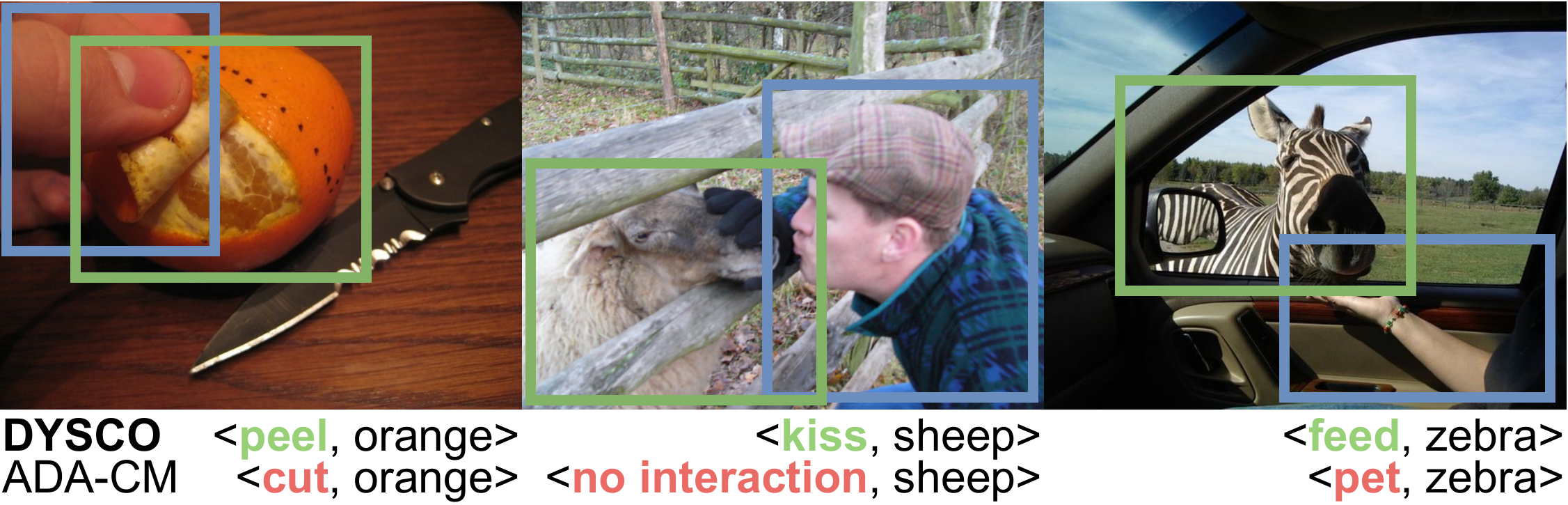}
\vspace{-2em}
\caption{\textbf{Qualitative results of our \method (top)} and ADA-CM~\cite{Lei2023} (bottom).}
\Description{Qualitative results of our \method and ADA-CM.}
\label{fig:qualitatives}
\end{figure}

Fig.~\ref{fig:qualitatives} shows the interactions predicted from both our \method and ADA-CM~\cite{Lei2023} on some samples of the HICO-DET dataset. Our predictions consistently outperform those of ADA-CM~\cite{Lei2023}, aligning with the quantitative results. Please refer to the Supp. Mat.) for additional examples.
\begin{figure}[!ht]
\centering
\vspace{-0.5em}
\includegraphics[width=0.95\linewidth]{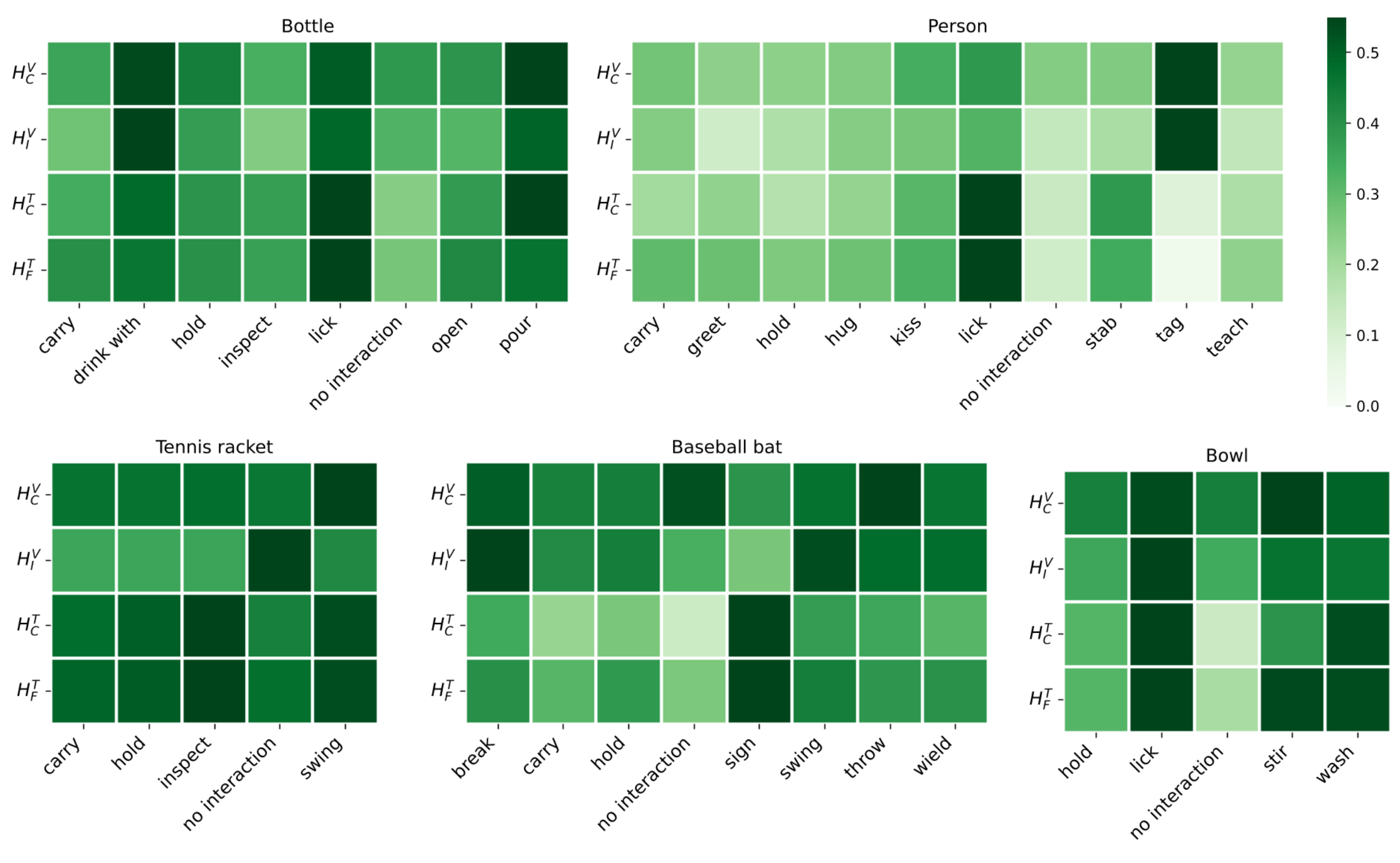}
\vspace{-1.5em}
\caption{\textbf{Behavior of our \multihead} applied to objects and verbs of the HICO-DET dataset.}
\Description{Behavior results of our \multihead.}
\label{fig:qualitatives_mhom}
\vspace{-1.5em}
\end{figure}
\cref{fig:qualitatives_mhom} shows how our \multihead dynamically adjusts each head's contribution to different verbs and objects of the HICO-DET dataset.
For example, in the case of the object \texttt{``bottle''}, the verb \texttt{``drink with''} gets the most attention from the visual heads, while \texttt{``lick''} relies on all heads, and \texttt{``pouring''} favors contextual visual and coarse text heads.
This highlights the importance of the \multihead module and its effectiveness for HOI.
    
\section{Conclusions} \label{sec:conclusions}
We have presented \method, a novel, training-free approach for HOI detection that advances SOTA performances by effectively combining textual and visual cues. Our method introduces innovative interaction signatures to improve the semantic alignment between interaction representations and visual features.
We also cast the HOI detection task as a multi-head attention process, enabling the dynamic reweighting of multimodal features, a unique contribution to the field. This dynamic reweighting allows our method to adapt to varying contributions of visual and textual features, which is a significant improvement over previous work that relied on fixed probability weighting.
There remains room to enhance our registry and the quality of feature representations, particularly by improving the text encoder's comprehension of verbs. The object detector (which is used consistent with the existing studies for fair comparisons) also impacts performance and could be improved in future developments.

\clearpage
\section*{Acknowledgements}
We thank CINECA and the ISCRA initiative for the availability of high-performance computing resources. 
This work was supported by the EU Horizon ELIAS (No. 101120237), IAMI (No. 101168272), and ELLIOT (No. 101214398) projects, by the MUR PNRR project FAIR - Future AI Research (PE00000013), and by the PRIN B-FAIR (Prot. 2022EX F3HX) project.
This work was carried out in the Vision and Learning joint laboratory of FBK and UNITN.

\bibliographystyle{ACM-Reference-Format}
\balance
\bibliography{main}

\clearpage
\setcounter{page}{1}
\maketitlesupplementary

The supplementary material provides a deeper exploration of \method, our novel training-free approach that utilizes rich textual and visual information for robust HOI detection.
Specifically, we include additional implementation details (\cref{sec:supp-implementation}), 
provide additional experiments on the registry size (\cref{sec:supp-registry}),
present further experiments on label-free HOI (\cref{sec:supp_pseudolabels}), as well as visualizations of our interaction signatures and \method's outputs (\cref{sec:supp-qualitatives}).

\section{Additional implementation details} \label{sec:supp-implementation}
The object detector, DETR~\cite{Carion2020}, is built upon a ResNet-50 backbone for feature extraction and leverages an encoder-decoder architecture to predict object bounding boxes and their corresponding labels.
For the visual~$\phi^{v}$ and textual~$\phi^{t}$ encoders, we use the pretrained checkpoints provided by OpenAI~\cite{Radford2021} for the ViT-B/16 and ViT-L/14 backbones.
Additionally, for LongCLIP~\cite{Zhang2024}, we employ the official checkpoints made available by the authors.

\section{Registry size} \label{sec:supp-registry}
\begin{figure}[H]
    \centering
    \includegraphics[width=1.0\linewidth]{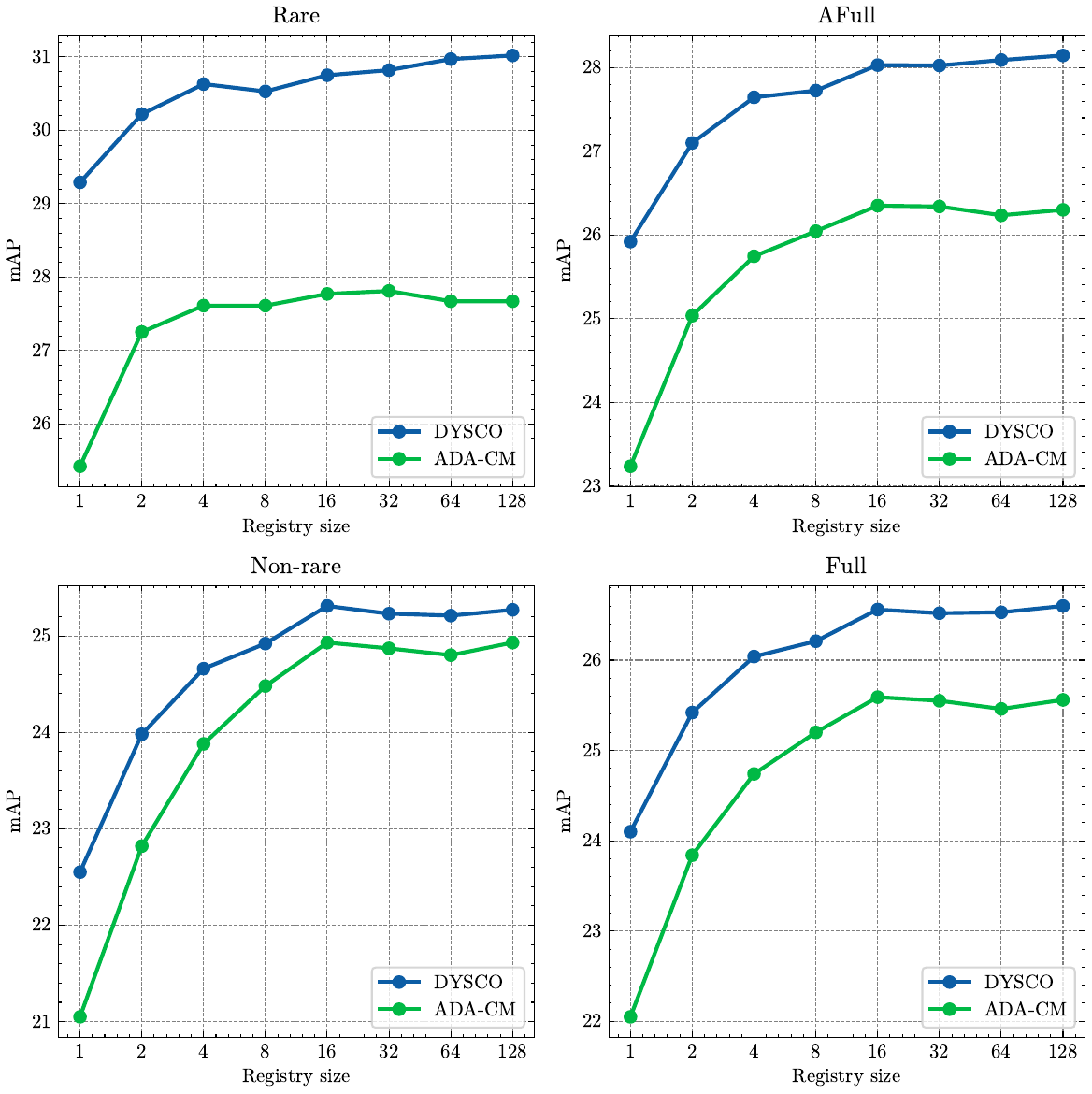}
    \caption{\textbf{Effect of the registry $\bm{\mathcal{R}}$ size} on \method. Tested on HICO-DET~\cite{Gupta2015} using ViT-B/16 as backbone.}
    \Description{Effect of the registry size on \method.}
    \label{fig:supp_abl_registry_size}
\end{figure}

We investigate how the number of visual samples $J$ in the registry $\mathcal{R}$ affects the performance of \method, as shown in \cref{fig:supp_abl_registry_size}.
To evaluate this, we explore different values $J \in \{1, 2, 4, 8, 16, 32, 64, 128\}$, conducting the experiments on the HICO-DET dataset~\cite{Gupta2015}, using ViT-B/16 as the backbone architecture.

The results indicate that performance improves steadily as the registry size increases, peaking at $J = 128$ in terms of overall accuracy (\ie, ``Full'' results).
Notably, unlike ADA-CM~\cite{Lei2023}, where $J \geq 16$ negatively impacts the performance, our \method performance improves as we increase $J$.
Nonetheless, we select $J = 8$ as the default registry for all experiments in the paper, in line with prior art~\cite{Lei2023}.

\section{Additional experiments on label-free HOI}
\label{sec:supp_pseudolabels}
Following~\cref{sec:pseudolabels}, we present supplementary experiments evaluating \methodaf under varying confidence thresholds. As shown in \cref{tab:supp_pseudo_clip}, \methodaf achieves performance comparable to \method across different probability thresholds $\bm{p}$, with optimal results obtained when constructing the registry $\mathbb{R}$ with increasing confidence ($\bm{p} \geq 0.9$).
We further explore employing multimodal large language models (MLLMs) like LLaVA~\cite{li2024llava} for pseudolabel extraction. In such setting, interaction scores are estimated measuring the likelihood of replying positively to a simple question, \ie $P(\text{``Yes''} \mid \text{``Is the person \texttt{\{verb\}} the \texttt{\{object\}}?''})$. Such likelihood represents the MLLM's prediction for each HOI instance. \Cref{tab:supp_pseudo_llava} demonstrates that \methodaf maintains competitive performance even with MLLM-generated pseudolabels, paving the way for advancing weakly-supervised training-free HOI detection frameworks.

\begin{table}[!ht]
\centering
\caption{\textbf{Performance of \methodaf} at different thresholds $\mathbf{p}$. The best results are marked in \textbf{bold}.}
\resizebox{0.78\columnwidth}{!}{%
\begin{tabular}{@{}lcccc@{}}
\toprule
\textbf{$\bm{p}$} & \textbf{Rare} & \textbf{Non-rare} & \textbf{AFull} & \textbf{Full} \\ \midrule
0.5                                          & 28.67\%	& 22.93\%	& 25.80\%	& 24.25\%       \\
0.6                                          
& 29.23\%	& 22.93\%	& 26.08\%	& 24.38\%       \\
0.7                                          & 29.07\%	& 22.96\%	& 26.01\%	& 24.36\%       \\
0.8                                          & 29.18\%	& 22.93\%	& 26.06\%	& 24.37\%       \\
0.9                                         & \textbf{29.29\%}	& \textbf{22.92\%}	& \textbf{26.10\%}	& \textbf{24.38\%}             \\ \bottomrule
\end{tabular}%
}
\label{tab:supp_pseudo_clip}
\end{table}

\begin{table}[!ht]
\caption{Performace results of our \methodaf with LLaVA~\cite{li2024llava} as our model for pseudolabeling. LF stands for ``label-free''. The best results for each setting are marked in bold.}
\resizebox{0.9\columnwidth}{!}{%
\begin{tabular}{@{}lccccc@{}}
\toprule
\multicolumn{1}{c}{\textbf{Method}} & \textbf{LF} & \textbf{Rare}   & \textbf{Non-rare} & \textbf{AFull}  & \textbf{Full}   \\ 
\midrule
ADA-CM$\dagger$~\cite{Lei2023} & \xmark & 27.61\% & 24.48\% & 26.04\% & 25.20\% \\
\method (Ours) & \xmark & \textbf{30.53\%} & \textbf{24.92\%}   & \textbf{27.73\%} & \textbf{25.75\%} \\
\midrule
LLaVA OV 7B~\cite{li2024llava} & \cmark   & 27.34\% & 20.03\%   & 23.68\% & 21.71\% \\
\methodaf (Ours) & \cmark & \textbf{29.63\%} & \textbf{23.92\%}   & \textbf{26.78\%} & \textbf{25.23\%} \\
\bottomrule
\end{tabular}%
}
\label{tab:supp_pseudo_llava}
\end{table}

\section{Additional visualizations} \label{sec:supp-qualitatives}
\begin{figure*}[!t]
\centering
\subfloat[]{\includegraphics[width=0.71\linewidth]{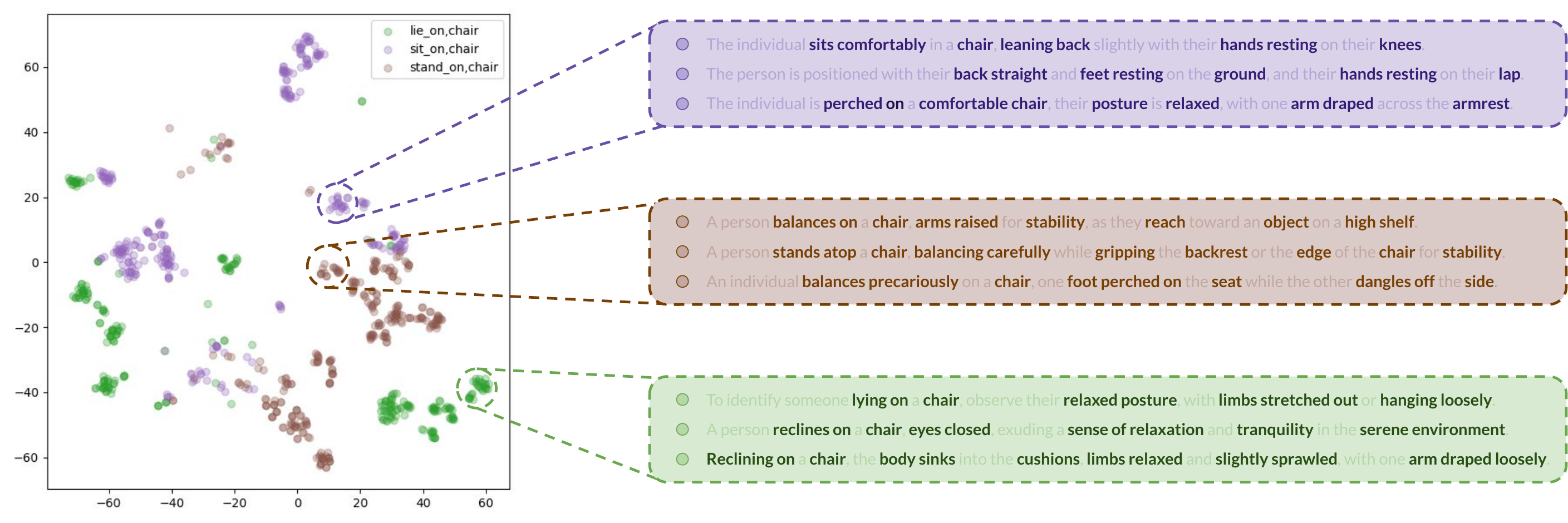}
  \label{fig:supp-tsne-00}}
\\
  \subfloat[]{\includegraphics[width=0.25\linewidth]{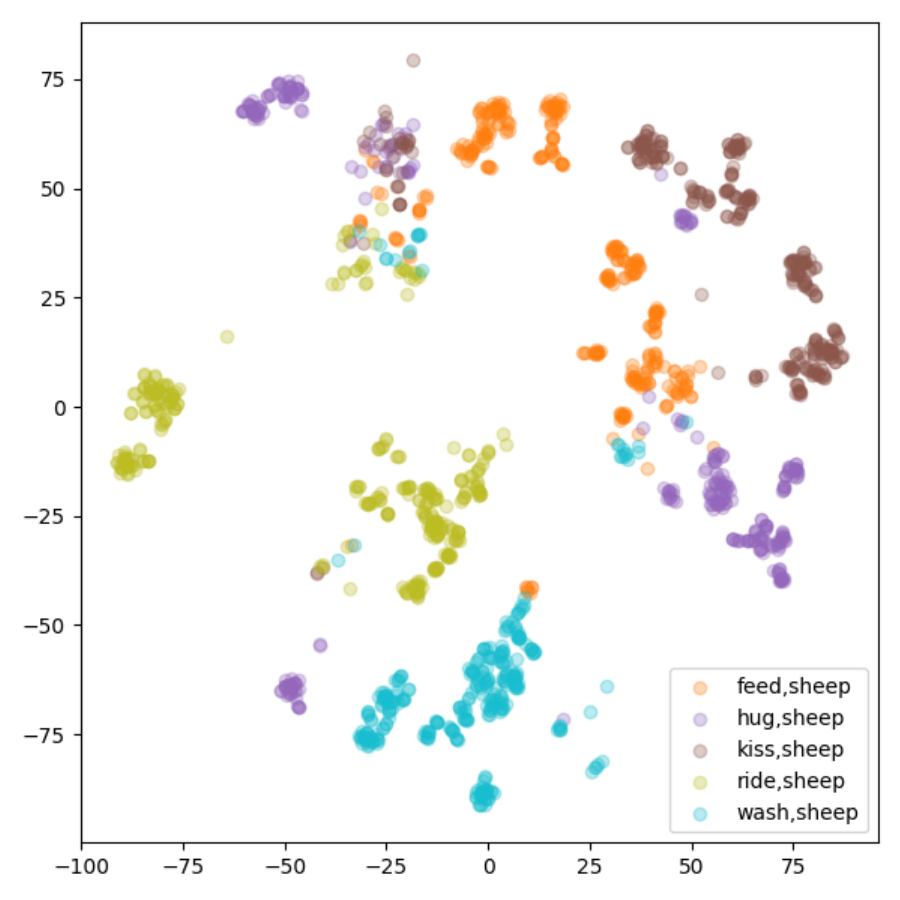}
  \label{fig:supp-tsne-01}}
\hfill
  \subfloat[]{\includegraphics[width=0.25\linewidth]{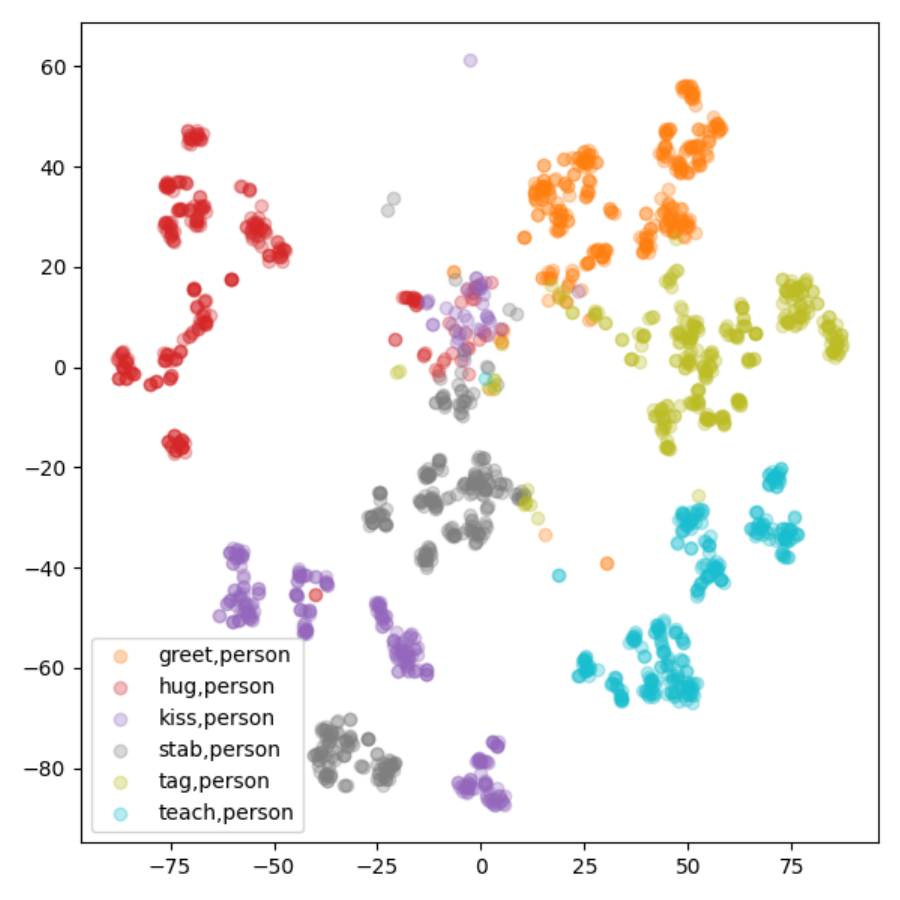}
  \label{fig:supp-tsne-02}}
\hfill
  \subfloat[]{\includegraphics[width=0.25\linewidth]{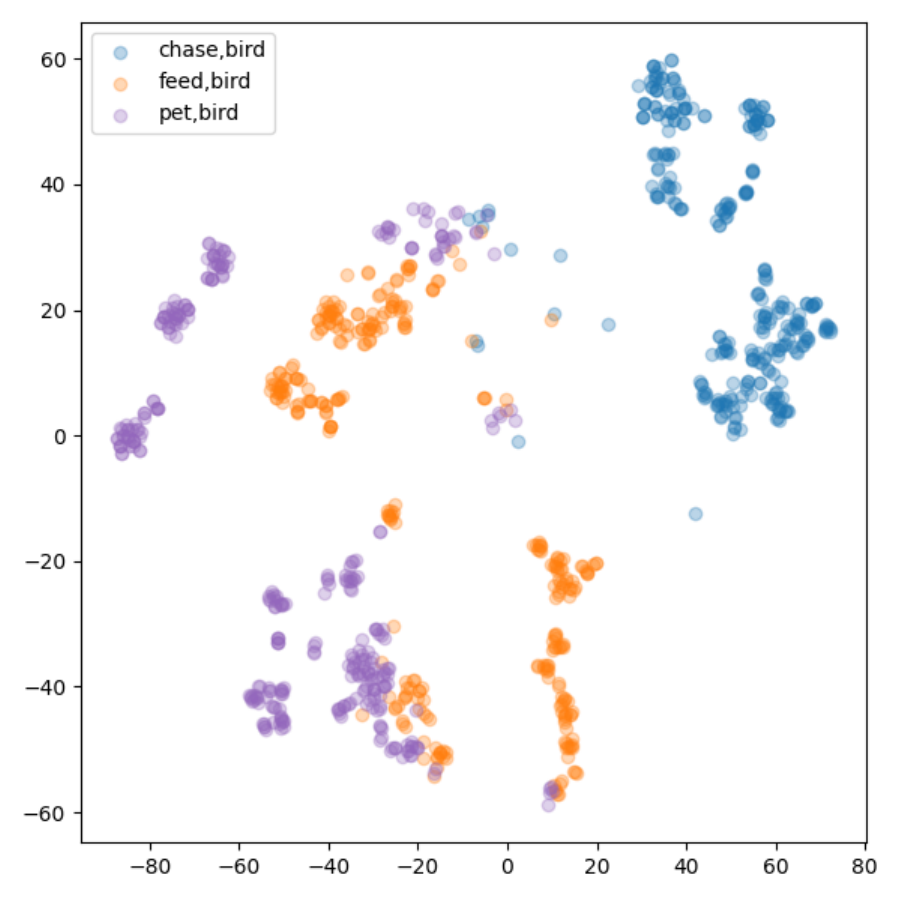}
  \label{fig:supp-tsne-03}}
\\
  \subfloat[]{\includegraphics[width=0.25\linewidth]{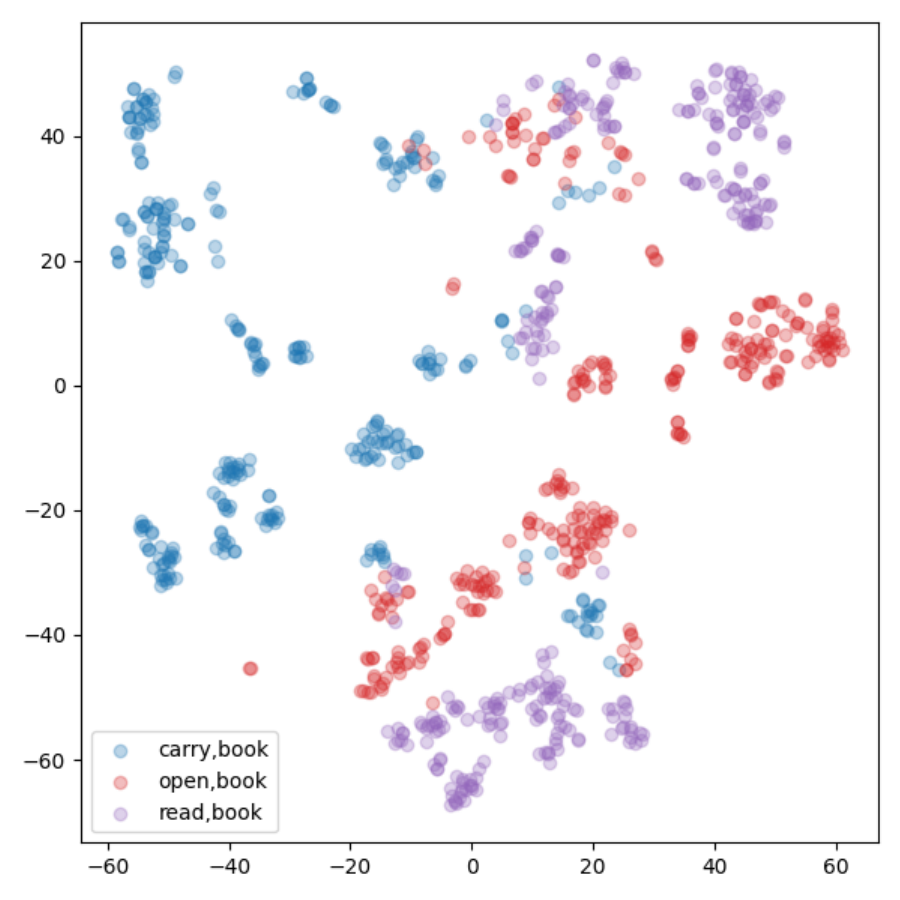}
  \label{fig:supp-tsne-04}}
\hfill
  \subfloat[]{\includegraphics[width=0.25\linewidth]{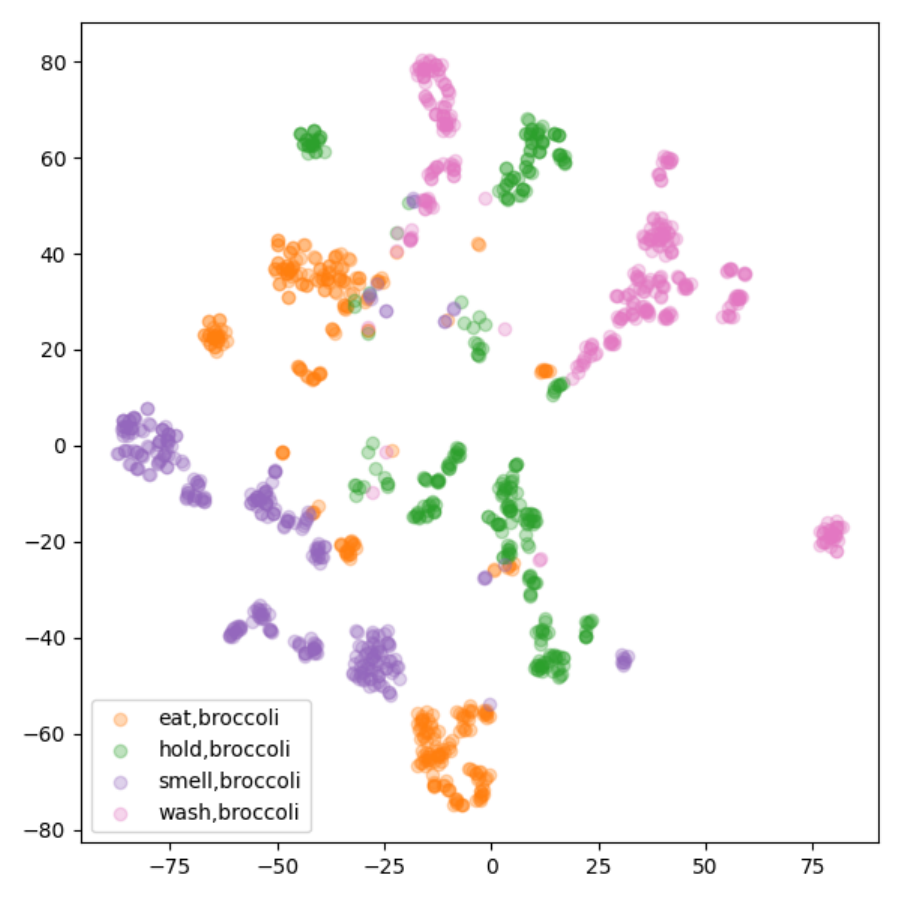}
  \label{fig:supp-tsne-05}}
\hfill
  \subfloat[]{\includegraphics[width=0.25\linewidth]{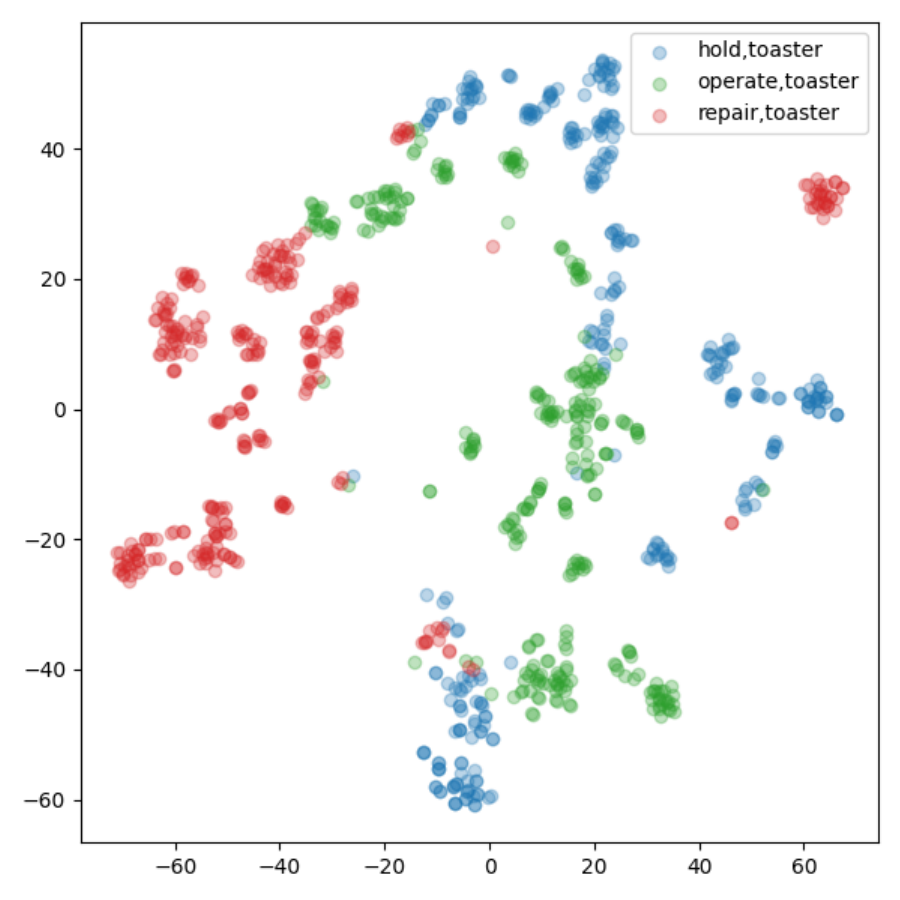}
  \label{fig:supp-tsne-06}}
\\
  \subfloat[]{\includegraphics[width=0.25\linewidth]{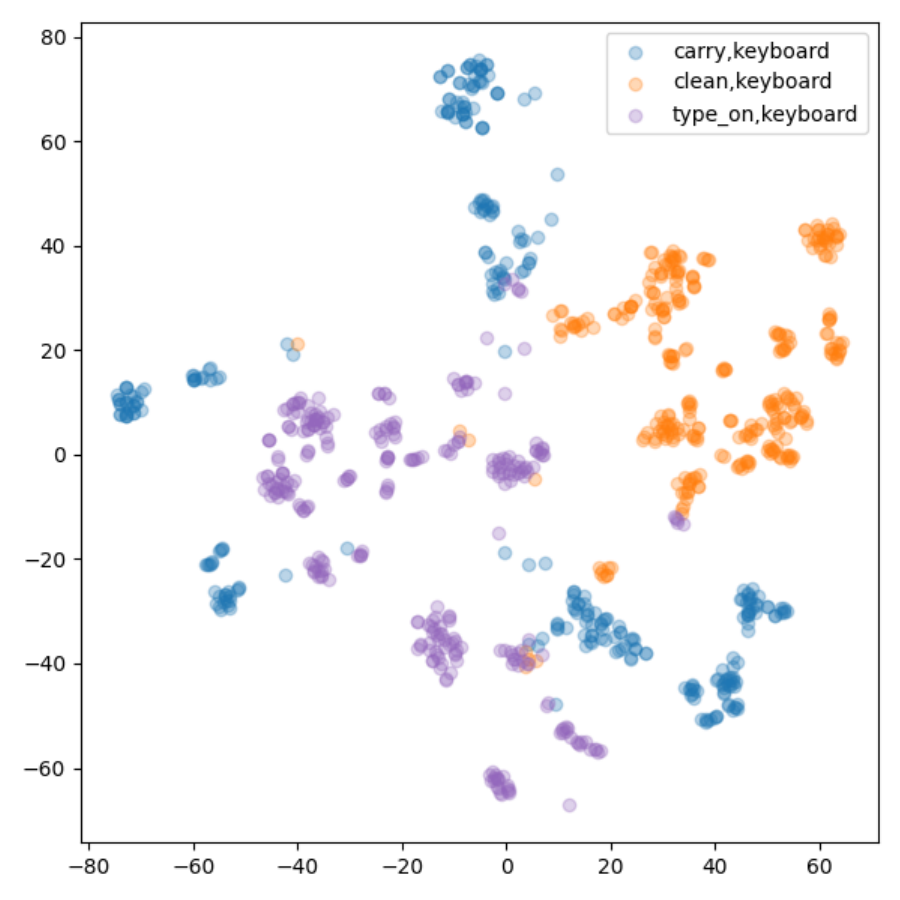}
  \label{fig:supp-tsne-07}}
\hfill
  \subfloat[]{\includegraphics[width=0.25\linewidth]{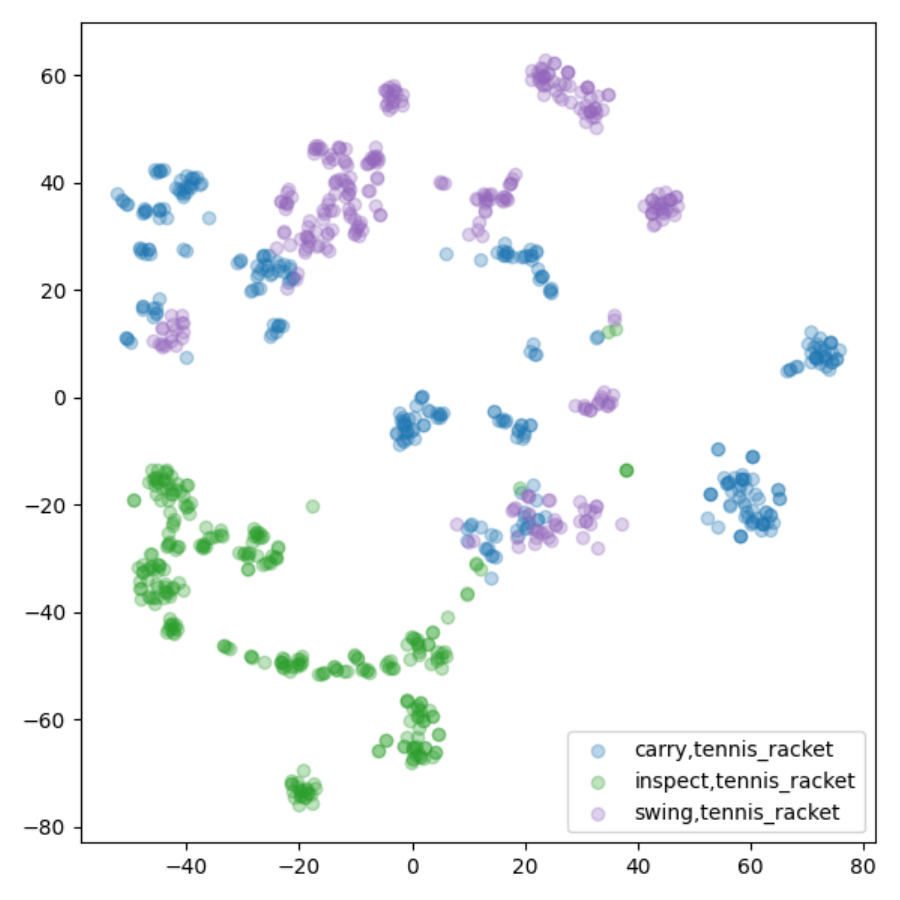}
  \label{fig:supp-tsne-08}}
\hfill
  \subfloat[]{\includegraphics[width=0.25\linewidth]{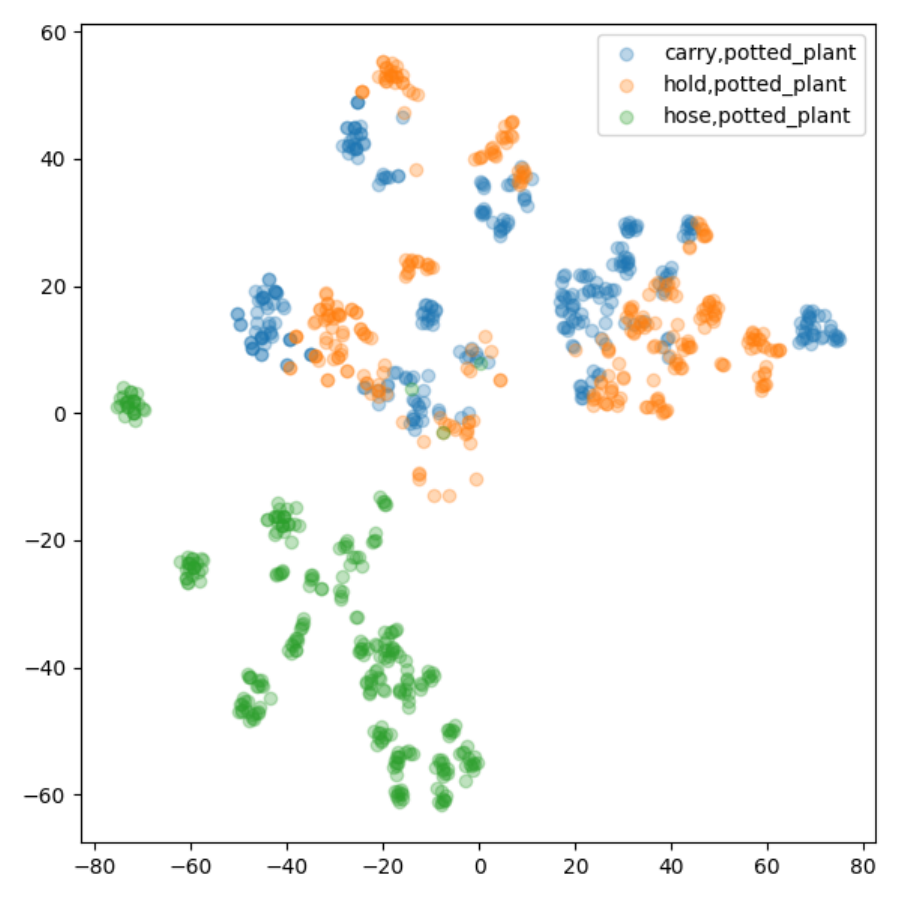}
  \label{fig:supp-tsne-09}}

\caption{\textbf{T-SNE representations of interaction signatures} for the objects \protect\subref{fig:supp-tsne-00}~chair, \protect\subref{fig:supp-tsne-01}~sheep, \protect\subref{fig:supp-tsne-02}~person, \protect\subref{fig:supp-tsne-03}~bird, \protect\subref{fig:supp-tsne-04}~book, \protect\subref{fig:supp-tsne-05}~broccoli, \protect\subref{fig:supp-tsne-06}~toaster, \protect\subref{fig:supp-tsne-07}~keyboard, \protect\subref{fig:supp-tsne-08}~tennis racket, and \protect\subref{fig:supp-tsne-09}~potted plant. 
}
\Description{T-SNE representations of interaction signatures for different objects.}
\label{fig:supp-tsne}
\vspace{-5em}
\end{figure*}

\cref{fig:supp-tsne} illustrates the interaction signature representations for some objects found in the HICO-DET dataset~\cite{Gupta2015}.
The flexibility of our method effectively captures the intricate structure of complex, stratified manifolds.
Consequently, interaction signatures for semantically related concepts that frequently co-occur (\eg, \texttt{``hold''} and \texttt{``carry''} in \cref{fig:supp-tsne-09}) are positioned in close proximity and exhibit similar patterns.
In contrast, interactions that are conceptually distinct (\eg, \texttt{``hug''} and \texttt{``teach''} in \cref{fig:supp-tsne-02}) are clearly separable, demonstrating the robustness of our approach in distinguishing interaction types.

We also provide the textual representation of our interaction signatures in \cref{fig:supp-tsne-00} and \cref{tab:supp_signatures_text}, demonstrating that they are closely linked to the source interaction and capture highly semantic details, such as related objects and attributes.

\subsection{Qualitative results}
\begin{figure*}[!ht]
\centering
\includegraphics[width=0.645\linewidth]{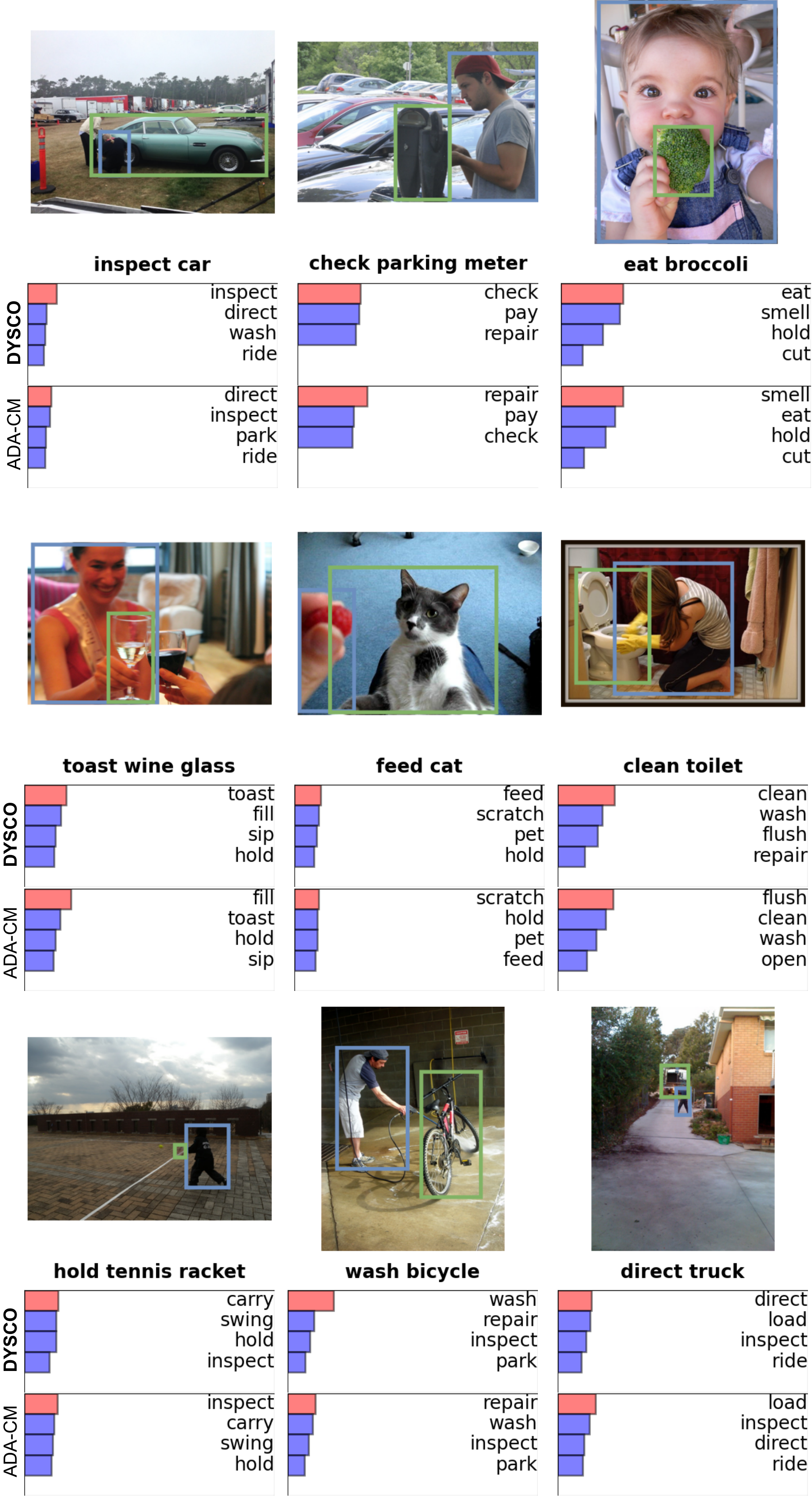}
\caption{\textbf{Qualitative results of our \method (top)} and ADA-CM~\cite{Lei2023} (bottom). \textbf{Bold} is ground-truth, while \textcolor{red}{red} bar is the top-1 prediction.}
\Description{Additional qualitative results of our \method and ADA-CM.}
\label{fig:supp_qualitatives_best}
\end{figure*}

\begin{figure*}[!ht]
\centering
\includegraphics[width=0.645\linewidth]{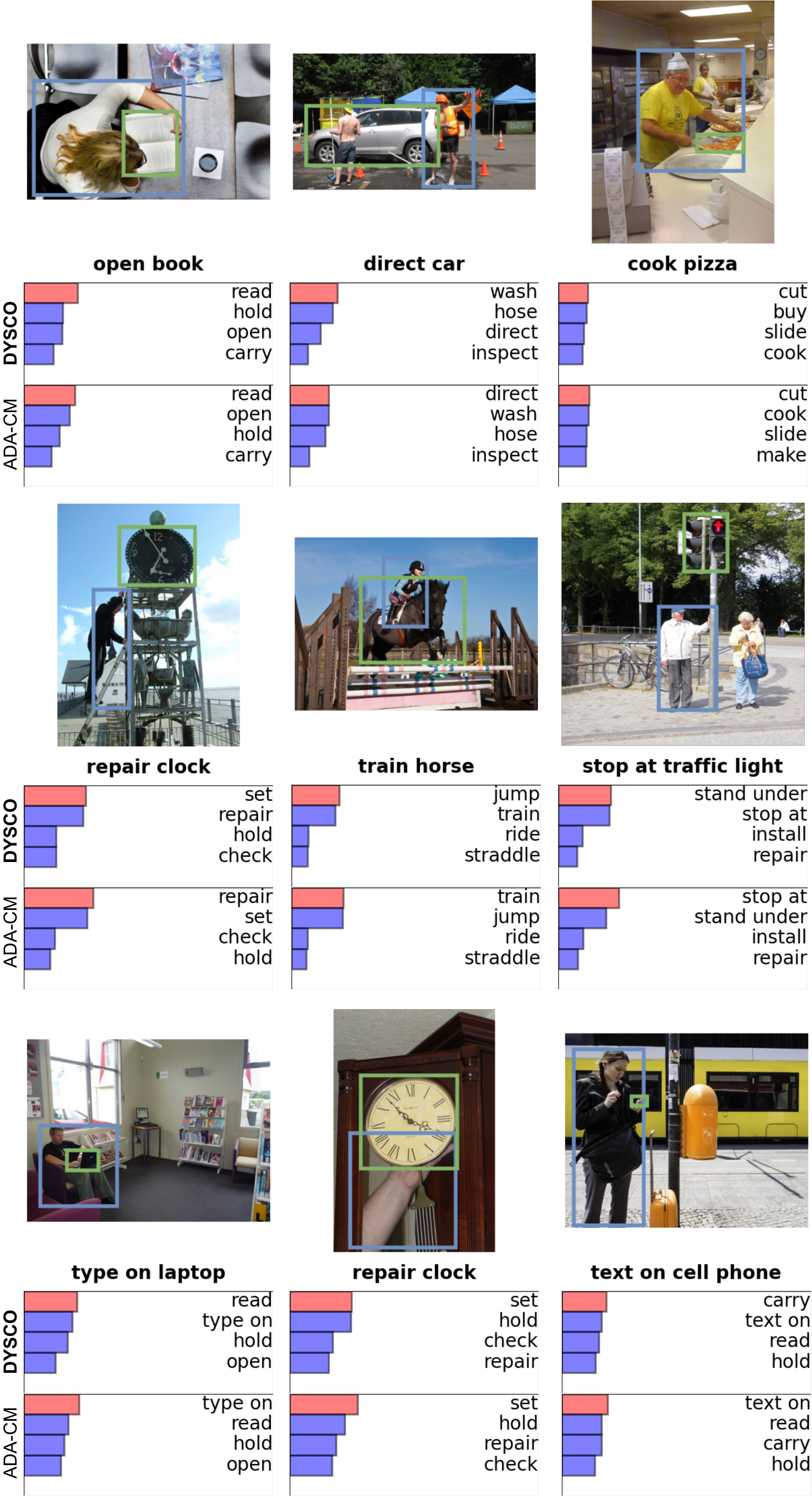}
\caption{\textbf{Failure cases of our \method (top)} and ADA-CM~\cite{Lei2023} (bottom). \textbf{Bold} is ground-truth, while \textcolor{red}{red} bar is the top-1 prediction.}
\Description{Qualitative results of our \method and ADA-CM showcasing failure cases.}
\label{fig:supp_qualitatives_bad}
\vspace{-5em}
\end{figure*}

We provide additional qualitative results in \cref{fig:supp_qualitatives_best,fig:supp_qualitatives_bad}, showcasing the predictions made by \method compared to those of ADA-CM~\cite{Lei2023}.

The first set of examples in \cref{fig:supp_qualitatives_best} demonstrates the effectiveness of our method, particularly in images where subtle cues are crucial for understanding the interaction.
For instance, in the cases of \texttt{``checking a parking meter''} and \texttt{``washing a bicycle''}, our model excels at capturing these nuances.
These results highlight the benefits of our multi-head attention mechanism, which effectively integrates both fine-grained and coarse-grained information.

On the other hand, \cref{fig:supp_qualitatives_bad} shows some failure cases of our model.
However, it is important to note that, in some instances, our model's predictions seem more fitting than the ground-truth labels.
For example, in the case of \texttt{``training a horse''} (center image), where our model predicts \texttt{``jumping with a horse''}; \texttt{``repairing a clock''} (center-bottom image), where the prediction is \texttt{``setting a clock''}; and \texttt{``opening a book''} (top-left image), where it predicts \texttt{``reading a book''}.
In these instances, our model’s predictions appear even more plausible than the ground-truth annotations, still underlining \method's potential for providing a deep understanding about HOI in a wide variety of contexts.

\begin{table*}[!htbp]
\begin{tcolorbox}[breakable, enhanced jigsaw,title={feed, sheep}]
``Gently kneeling beside the sheep, the person holds out a handful of feed, fingers splayed. The sheep approaches eagerly, nuzzling the offering with its wet nose, while the person quietly observes, fostering a connection through this simple act of nourishment.''

``The person crouches down, gently extending their hand filled with feed towards the sheep, who eagerly approaches, nibbling at the food. With each bite, the soft sound of munching fills the air, underscored by the gentle rustle of grass.''
\end{tcolorbox}

\begin{tcolorbox}[breakable, enhanced jigsaw,title={hug, person}]
``A person extends their arms wide, approachingly enveloping another in an embrace. As they pull each other close, there is a moment of stillness, often accompanied by gentle swaying, conveying warmth and connection through shared energy and comfort.''

``As two individuals lean towards each other, their arms encircle one another in a warm embrace, conveying comfort and connection. They often close their eyes or lean in, reflecting a moment of shared intimacy and solace.''
\end{tcolorbox}

\begin{tcolorbox}[breakable, enhanced jigsaw,title={chase, bird}]
``The individual sprints with determination, arms pumping at their sides, as their eyes track the bird flitting from branch to branch. With each quickened step, they navigate around obstacles, excitement evident in their movements, despite the birds effortless evasion.''

``With swift and agile movements, one lunges forward, arms outstretched, attempting to close the distance. The chase is punctuated by quick bursts of speed, as the bird flutters and darts away, weaving gracefully through the air, evading capture.''
\end{tcolorbox}

\begin{tcolorbox}[breakable, enhanced jigsaw,title={carry, book}]
``Carrying a book typically involves holding it securely, with fingers wrapped around the spine or cover, while keeping it close to the body to maintain balance. The individual may occasionally adjust their grip or shift the book to ensure comfort.''

``Carrying a book involves gripping it with one or both hands, often with the cover facing inward or outward, depending on the context. The individual may hold it close to the chest or at their side, occasionally pausing to read or adjust their grip.''
\end{tcolorbox}

\begin{tcolorbox}[breakable, enhanced jigsaw,title={eat, broccoli}]
``Grasping a florets of broccoli, one brings it to the mouth, its vibrant green surface glistening slightly. As the teeth sink into the firm texture, a gentle crunch resonates, releasing a fresh, earthy flavor that fills the palate, leaving a satisfying sensation.''

``The person carefully picks up a floret of broccoli, bringing it to their mouth. They take a bite, crunching through the tender yet firm texture, while releasing a vibrant green aroma, savoring the natural flavor before swallowing.''
\end{tcolorbox}

\begin{tcolorbox}[breakable, enhanced jigsaw,title={hold, toaster}]
``The hands grasp the sides of the toaster firmly, fingers curled around the edges while the body tilts slightly to maintain balance. The appliance is lifted close, revealing the gleaming metal finish and intricate slots designed for bread.''

``The person grips the toaster firmly, fingers wrapping around the sides for stability. Their thumb rests on the lever, poised to activate the mechanism, while the other hand may gently support the bottom to prevent any unsteady movements.''
\end{tcolorbox}
\caption{\textbf{Text representations for some of the interaction signatures} employed by \method on the HICO-DET~\cite{Gupta2015} dataset.}
\label{tab:supp_signatures_text}
\end{table*}

\end{document}